%% file: arxiv_main.tex
\documentclass[11pt]{article}

\usepackage[margin=1in]{geometry}
\usepackage{mathpazo}
\usepackage{natbib}
\usepackage{subcaption}
\usepackage[colorlinks,citecolor=blue]{hyperref}
\usepackage{bbm}
\usepackage[disable]{todonotes}
\usepackage{lineno}

\usepackage{inconsolata}
\usepackage{url}            
\usepackage{booktabs}       
\usepackage{amsfonts}       
\usepackage{nicefrac}       
\usepackage{microtype}      
\usepackage{xcolor}         
\usepackage{multicol}
\usepackage{enumitem}
\usepackage{psfrag}
\usepackage{verbatim}
\usepackage{mathrsfs}
\usepackage{amssymb}%
\usepackage{pifont}%
\usepackage{multirow}
\usepackage{longtable}
\usepackage{listings}
\usepackage{graphicx}
\usepackage{amsmath}
\usepackage{mathtools}
\usepackage{amsthm}
\usepackage{xspace}
\usepackage[subtle]{savetrees}
\usepackage[most]{tcolorbox}
\usepackage{colortbl}
\usepackage[capitalize,noabbrev]{cleveref}

\newcommand{\OURS}{\textit{Anthology}\xspace}
\newcommand{\BASEQA}{\texttt{QA}\xspace}
\newcommand{\BASEBIO}{\texttt{Bio}\xspace}

\newcommand{\GREEDY}{greedy\xspace}
\newcommand{\MAXSUM}{max weight\xspace}

\definecolor{berkeleyblue}{HTML}{3B7EA1}
\definecolor{berkeleygold}{HTML}{FDB515}
\definecolor{main}{HTML}{4472C4}   
\definecolor{sub}{HTML}{EBF4FF}  
\newcommand\cb{\cellcolor{berkeleyblue!20}}
\newcommand\cy{\cellcolor{berkeleygold!20}}

\DeclareMathSizes{10}{9}{6}{5}
\newcommand\blfootnote[1]{%
  \begingroup
  \renewcommand\thefootnote{}\footnote{#1}%
  \addtocounter{footnote}{-1}%
  \endgroup
}

\title{Virtual Personas for Language Models via\\an Anthology of Backstories}

\author{
Suhong Moon\blfootnote{Equal contribution
}$^{*}$ \enspace\enspace Marwa Abdulhai$^{*}$ \enspace\enspace Minwoo Kang$^{*}$ \enspace\enspace Joseph Suh$^{*}$\\
Widyadewi Soedarmadji\enspace\enspace Eran Kohen Behar\enspace\enspace David M. Chan
\\
{University of California, Berkeley}\\[6pt]
{John Canny}\\ 
{
University of California, Berkeley and Google LLC\blfootnote{No Google authors used or analyzed data from Llama models
}}\vspace{3mm}\\
    {{\small{suhong.moon}@berkeley.edu}}\\
\vspace{5mm}\\
 \color{red}{{\small\underline{Warning}: \emph{This paper includes examples and model-generated content that may be considered offensive.}}}\\ 
 \vspace{-5mm}\\
}
\date{}

\begin{document}

\maketitle

\input{sections/_s0_abstract}
\input{sections/_s1_introduction}
\input{sections/_s2_method}
\input{sections/_s3_experiments}
\input{sections/_s4_results}
\input{sections/_s5_related}

\input{sections/_s6_limitations}
\input{sections/_s7_conclusion}

\section*{Acknowledgements}
We appreciate the valuable feedback from John Canny and Sehoon Kim. We also thank Alia Braley and Alane Suhr for the fruitful discussions. S.M. and J.S. would like to acknowledge the support from the Korea Foundation for Advanced Studies (KFAS). Additionally, authors, as part of their affiliation with UC Berkeley, were supported in part by the National Science Foundation, US Department of Defense, and/or the Berkeley Artificial Intelligence Research (BAIR) industrial alliance program.

\bibliography{anthology}
\bibliographystyle{bib_style} %
\clearpage

\appendix
\input{sections/_a_chat_models}
\input{sections/_a_backstory_details_new}
\input{sections/_a_experiment_details}
\input{sections/_a_human_survey_details}

\input{sections/_a_demographic_survey}
\clearpage

\end{document}

%% file: sections/_s0_abstract.tex
\begin{abstract}
Large language models (LLMs) are trained from vast repositories of text authored by 100s of millions of distinct authors, reflecting an enormous diversity of human traits. 
While these models could potentially model a variety of human behaviors, most prior work has focused on predicting behaviors of {\em groups} of subjects, or have relied on 
a small set of short hand-written ``personas''. 
In this work, we introduce ``\OURS'', a method for conditioning LLMs to particular \emph{virtual personas} by harnessing open-ended life narratives, which we refer to as ``backstories.'' Another key difference from most previous work is our use of {\em pretrained} instead of instruction-tuned models, since instruction-tuning suppresses many natural human traits and most human diversity.
We show that our methodology enhances the consistency and reliability of experimental outcomes while ensuring better representation of diverse sub-populations.
Across three nationally representative human surveys conducted as part of Pew Research Center's American Trends Panel (ATP), we demonstrate that \OURS achieves up to 18\% improvement in matching the response distributions of human respondents and 27\% improvement in consistency metrics. Beyond quantitative studies, we believe such a pool of subjects can provide valuable {\em qualitative feedback} to an experimenter via open-ended interviews of the model conditioned on a backstory. 
\end{abstract}

%% file: sections/_s1_introduction.tex
\section{Introduction}
\label{section:intro}

Large language models (LLMs) are trained from vast repositories of human-written text~\citep{Touvron2023LLaMA,llama3,gpt3,gpt-4o,mixtral-8x22b,Jiang2024Mixtral}. 
These texts are authored by 100s of millions of distinct authors, reflecting an enormous diversity of human traits~\cite{choi2024beyond,wolf2024fundamental}. 
As a result, when a language model completes a prompt, the generated response implicitly encodes a mixture of voices from human authors that have produced the training text from which the completion has been extrapolated. 
However this natural diversity of ``voices'' in language models is at odds with the requirements of most applications of LLMs: a single, helpful agent voice, factual answers and a minimum of human-like emotion in responses. Much of the later tuning of LLMs (especially RLHF in chat models such as chatGPT and GPT 4) has been shown to reduce this diversity \cite{maxmin_rlhf}. It also clearly suppresses negative human attitudes and behaviors. Here we show that with careful design and use of ``upstream'' (before instruction tuning) pre-trained models, its possible to preserve compelling and realistic virtual human voices, and apply them to practical human simulation tasks. 

There is growing recent interest in the use of LLMs as human proxies for behavioral studies \cite{ANESpaper,cognitive_psy_gpt3,llm_opinions,discovering_lm_behaviors_with_modekl_written_evaluations,social_simulacra,moral_mimicry,personality_llm,political_ideology_chatbot,probing_partisan_llm,simulate_replicate_human_studies, abdulhai2023moral,park2024generativeagentsimulations1000}. While it is
premature and perhaps unrealistic to argue that LLMs can replace human studies, they do not have to to be
useful. In practice, most human studies involve a variety of compromises in scale, reach, representation and number of questions to be answered~\cite{ANESpaper}.
LLMs on the other hand, provide a low-cost, high-speed alternative that supports a nearly-infinite range of querying/conditioning over
target subjects. The pool of LLM voices (hundreds of millions) contains many under-represented voices, hard to access subjects (homeless, ill, disabled, incarcerated, non-cooperative) in a seemingly unlimited set of contexts. For the specific design presented here, LLM models are also highly {\em scrutable}. That is, subjects can be queried in natural language about {\em why} they behaved in a certain way; the ``study'' can be extended/modified in any way the experimenter choose, and the ``subjects'' will be available always. We believe the affordances of LLM human models are sufficiently different from human studies that they are best considered as a new kind of instrument for studying behavior, rather than a just a replacement or budget form of human studies. 

While there are evident risks from many uses of LLMs~\citep{bommasani2022opportunities,bai2022constitutional,hendrycks2023overview}, the use of language models as an adjunct/alternative to human studies can help experimenters satisfy best practices (Belmont Principles~\cite{belmont_report}) for human studies. They minimize harms since no human subjects are directly involved, and with careful design can improve representation (justice). 

\begin{figure}[!t]
    \centering
    \captionsetup{font=small}
    \includegraphics[width=0.95\linewidth]{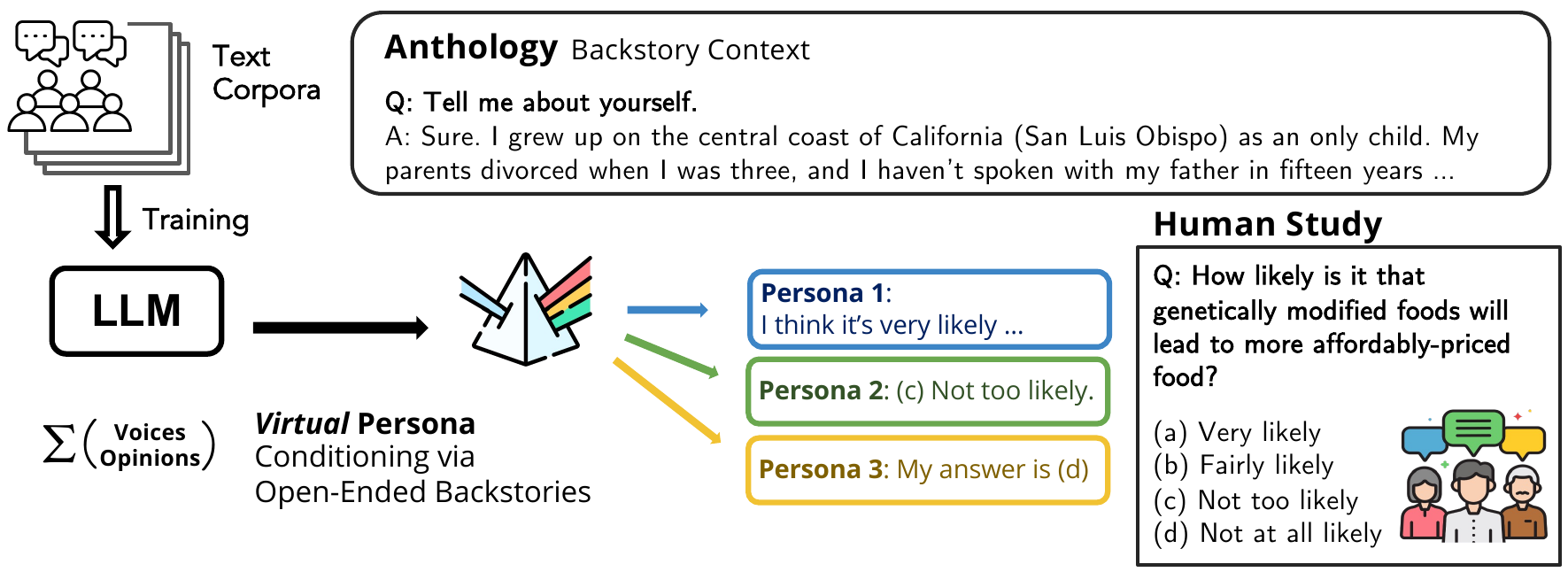}
    \caption{This work introduces \OURS, a method for conditioning LLMs to representative, consistent, and diverse virtual personas. We achieve this by generating naturalistic backstories, which can be used as conditioning context, and show that \OURS enables improved approximation of large-scale human studies compared to existing approaches in steering LLMs to represent individual human voices.}
    \label{fig:anthology_overview}
\end{figure}

For language models to effectively serve as virtual subjects, we must be able to steer their responses to reflect particular human users, \emph{i.e.} condition models to reliable \emph{virtual personas}.
To this end, existing work prompts LLMs with context that explicitly spell out the demographic and personal traits of the intended persona: for example, \cite{llm_opinions, liu2024evaluating, kim2024aiaugmented, hwang-etal-2023-aligning} attempt to steer LLM responses with a dialog consisting of a series of question-answer pairs about demographic indicators, a free-text biography listing all traits, and a portrayal of the said persona in second-person point-of-view.
While these approaches have shown modest success, they have been limited in (i) closely representing the responses of human counterparts, (ii) consistency, and (iii) successfully binding to diverse personas, especially those from under-represented sub-populations.

So how might we condition LLMs to virtual personas that are \emph{representative, consistent}, and\emph{ diverse}?  In this work, we investigate the use of naturalistic bodies of text describing individual life-stories, namely \emph{backstories}, as prefix to model prompts for persona conditioning. 
The intuition is that open-ended life narratives both explicitly and implicitly embody diverse details about the author, including age, gender, education level, emotion, and beliefs, etc.~\cite{mining_blogsphere, emotional_expression,personality_gender_age_open_vocab_approach,sns_depression_analysis,word_use_suicidal_poets}.
Lengthy backstories thus narrowly constrain the user characteristics, including latent traits as personality or mental health that are not solicited explicitly~\cite{personal_narratives, narraitive_construction_of_reality}, 
and strongly condition LLMs to diverse personas.

In particular, we explore a methodology to generate backstories from LLMs themselves, as a means to efficiently produce massive sets of subjects covering a wide range of human demographics---which we refer to as an \emph{Anthology} of backstories. 
We also introduce a method to sample backstories to match a desired distribution of human population. 
Our overall methodology is validated with experiments approximating well-documented large-scale human studies conducted as part of Pew Research Center's American Trends Panel (ATP) surveys.
We demonstrate that language models conditioned with LLM-generated backstories provide closer approximations of real human respondents in terms of matching survey response distributions and consistencies, compared to baseline methods.
Particularly, we show superior conditioning to personas reflecting users from under-represented groups, with improvements of up to 18\% in terms Wasserstein Distance and 27\% in consistency.

Our contributions are summarized as follows:
\vspace{-3mm}
\begin{itemize}[leftmargin=3.3mm]
\setlength\itemsep{2pt}

\item We introduce \OURS, which employs LLM-generated backstories to further condition LLM outputs, demonstrating that \OURS more accurately approximates human response distributions across three surveys covering various topics and diverse demographic sub-groups (Sections~\ref{subsection:main_result} and~\ref{subsection:ablation_underrepresented}).
\vspace{-1mm}
\item We describe a method for matching virtual subjects conditioned by backstories to target human populations. This approach significantly enhances the approximation of human response distributions (Section~\ref{subsection:ablation_matching}).
\vspace{-1mm}
\item We provide an open-source anthology of approximately 10,000 backstories for future research and applications in a broad spectrum of human behavioral studies. Additionally, we make the code for producing, processing, and administering surveys publicly available.
\end{itemize}

\begin{figure*}[!t]
    \centering
    \captionsetup{font=small}
    \includegraphics[width=0.95\linewidth]{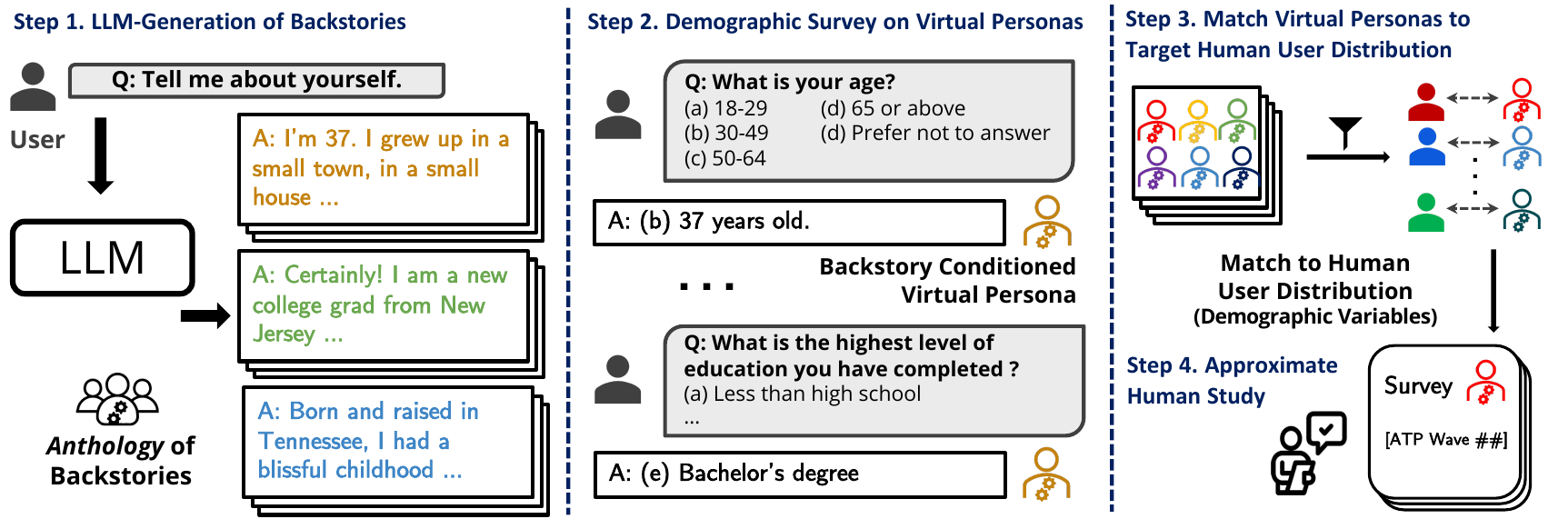}
    \caption{Step-by-step process of the \OURS approach which operates in four stages. First, we leverage a language model to generate an anthology of backstories using an unrestrictive prompt. Next, we perform demographic surveys on each of these backstory-conditioned personas to estimate the persona demographics. Following this, we methodologically select a representative set of virtual personas that match a desired distribution of demographics, based on which we administer the survey. We find that our approach can closely approximate human results (see Section~\ref{section:results} for details).}
    \label{fig:anthology_process}
\end{figure*}

%% file: sections/_s2_method.tex
\section{Conditioning LLMs to Virtual Personas via an \emph{\OURS} of Backstories}
\label{section:backstory_method}
In this section, we discuss details of the proposed \OURS approach.
We start with answering the core question: What are backstories and how might they help condition LLMs to particular personas when given as context?
With an example, we examine and lay out the advantages of conditioning models with backstories in Section~\ref{subsection:method_what_are_backstories}.

There are two practical considerations when using backstories as conditioned virtual personas for approximating human subjects. In the following sections, we discuss how we address each of these implications:
(i) We must acquire a substantial set of backstories that reflects a sufficient variety of human authors, since the target human study may require arbitrary demographic distribution of subjects. To this end, we introduce LLM-generated backstories to efficiently generate diverse backstories (Section~\ref{subsection:method_what_are_backstories}); and (ii) We cannot \emph{a priori} determine the possible demographic profile of a given backstory, since demographic variables may  not be explicitly mentioned in a naturalistic life narrative. Hence, we introduce methods to estimate demographics of the virtual persona conditioned by each backstory (Section~\ref{subsection:demographic_survey}) and sample subsets of backstories from anthology that match target human populations (Section~\ref{subsection:matching_population}). 

\subsection{What are Backstories?}
\label{subsection:method_what_are_backstories}
\input{figures/backstory_example_short}

We use the term \emph{backstories} to refer to first-person narratives that encompass various aspects of an individual's life, from where and how they grew up, their formative experiences, education, career, and personal relationships, to their values and beliefs.
These stories are inherently open-ended and personal, touching upon diverse facets of the author's demographic and personality traits. 

Consider the example shown in Figure~\ref{box:llm_generated_backstory_example_main}. 
We observe that the life story both \emph{explicitly} and \emph{implicitly} encodes information about the author, thereby providing rich insight into who the author is.
For instance, the backstory provides explicit hints about the author's age (``in my 60s''), hometown and/or region (``backwoods of this country''), and financial status during childhood (``grew up with very little''). 
But rather than being a simple listing of the aforementioned traits, the story itself embodies a natural, authentic voice of a particular human that reflects their values and personality. ~\cite{personal_narratives, narraitive_construction_of_reality}.

Our proposed approach is to condition language models with backstories by placing them as a prefixes to the LLM~\cite{gpt3, Touvron2023LLaMA} so as to strongly condition the ensuing text completion, in the same spirit of standard prompting approaches. 
As we see in Figure~\ref{box:llm_generated_backstory_example_main}, backstories capture a wide range of attributes about the author through high levels of detail and are naturalistic narratives that provide realism and consistency of the persona to which the LLM is conditioned. 

\subsection{LLM-Generated Backstories}
\label{subsection:method1_llm_generated_backstories}

A collection of human-written backstories could be drawn from existing sets of autobiographies or oral history collections.
The challenge, however, is both in terms of scale and diversity~\cite{yang-etal-2023-doc,yang2022re3}.
We find that, in their current standing, publicly available sources of autobiographical life narratives and oral histories are limited in the number of samples to sufficiently approximate larger human studies.

Custom human oral histories for LLM personas were recently collected in~\cite{park2024generativeagentsimulations1000}. This is a promising alternative approach, but is expensive and there are privacy challenges with distribution of such stories for living persons. 

Instead, we explore generation of backstories with pre-trained language models as a more scalable and cost-efficient alternative. We can also sample with finer control: e.g. tailor demographics to a particular study and/or over-sample minoritized groups to improve sample density and accuracy for those groups. 
As shown in Step 1 of Figure~\ref{fig:anthology_process}, we prompt LLMs with an open-ended prompt such as, ``Tell me about yourself.''
We specifically design the prompt to be simple so that the model responses as broad as possible (complex or academic language biases responses toward more highly-educated personas).

We believe that {\em generation} of plausible backstories as well as {\em accurate subsequent querying} of personas conditioned on those backstories requires the same capabilities in the language model. That is, if the model can accurately generate responses conditioned on a backstory and query, it should be able to {\em extend a partial backstory}, and thereby iteratively generate an entire backstory. See Figure ~\ref{box:llm_generated_backstory_example_main}.
With sampling temperature $T=1.0$, we generate backstories that encapsulate a broad range of life experiences of diverse human users. 
Further details about LLM generation of backstories, including examples, are summarized in Appendix~\ref{appendix:backstories}.

In our experience, instruction-tuned models (i.e. most LLM agent models) are completely unsuitable for this task. Whereas pre-trained models naturally represent an enormous spectrum of real users voices, 
instruction-tuned models have been trained towards a single helpful, largely unemotional voice. Attempting to prompt an 
instruction-tuned model for a backstory leads to short, evasive and vague answers. And queries to an instruction-tuned model
conditioned on a (real or synthetic) backstory lead to actions which are only positive and helpful, and avoiding the
(realistically human) actions that are not. 
Reinforcing this view, a recent paper~\cite{kapania2024simulacrumstoriesexamininglarge} studies the use of backstory-conditioned LLMs for qualitative (open-ended questioning) studies. The experimenters found a variety of disparities between the LLM responses and human responses. We argue that most of these disparities were due to the use of an instruction-tuned model "working as intended", but are not properties of LLMs more generally.

The good news is that while instruction-tuned models are far more widely used and 
available than pretrained models, all instruction-tuned models evolve from pretrained models. So access to pre-trained models is simply a matter of preserving earlier model snapshots before the 
instruction-tuning process begins. 

\subsection{Demographic Survey on Virtual Personas}
\label{subsection:demographic_survey}
As we intend to utilize virtual personas in the context of approximating human respondents in behavioral studies, it is critical that we curate an appropriate set of backstories that would condition personas representing the target human population. 
Each study would have a specific set of demographic variables and an estimation or accurate statistics of the demographics of its respondents.
Naturalistic backstories, despite their rich details about the individual authors, are however not guaranteed to explicitly mention all demographic variables of interest. 
Therefore, we emulate the process of how the demographic traits of human respondents have been collected---performing demographic surveys on virtual personas, as shown in Step 2 of Figure~\ref{fig:anthology_process}.

While we use the same set of demographic questions as used in the human studies, we consider that, unlike human respondents who each have a well-defined, deterministic set of traits, LLM virtual personas should be described with a \emph{probabilistic} distribution of demographic variables.
As such, we sample multiple responses for each demographic question to estimate the distribution of traits for the given virtual persona.
Further details about the process and prompts used in demographic surveys are described in Appendix~\ref{appendix:demographic_survey}.

\subsection{Matching Target Human Populations}
\label{subsection:matching_population}
The remaining question is: How do we choose the right set of backstories for each survey to approximate?
With the results of the demographic survey, we match virtual personas to the real human population, presented as Step 3 in Figure~\ref{fig:anthology_process}.
In doing so, we construct a complete weighted bipartite graph defined by the tuple, $G=(H, \ V, \  E)$.

The vertex set $H=\{h_1, \ h_2, \ \dots, \ h_n\}$ represents the human user group with the size of $n$,
while the other vertex set $V=\{v_1, \ v_2, \ \dots, \ v_m\}$ represents the virtual user group with the size of $m$.
Each vertex $h_i$ consists of demographic traits of $i-$th human user. 
Specifically, $h_i = (t_{i1},  \ t_{i2}, \ \dots,  \ t_{ik})$ where $k$ is the number of demographic variables, 
and $t_{il}$ is the $l-$th demographic variable's trait of $i-$th user.
Similarly, for each vertex in $V$, $v_j$ comprises probability distributions of demographic variables of each virtual user, defined as $v_j = \big( P(d_{j1}), \ P(d_{j2}), \ \dots, \ P(d_{jk})\big)$,
where $d_{jl}$ is $j-$th user's $l-$th demographic random variable and $P(d_{jl})$ is its probability distribution.

The edge set comprises $e_{ij} \in E$ which denotes the edge between $h_i$ and $v_j$.
The weight of an edge, $w(e_{ij})$ or equivalently $w(h_i, \ v_j)$, is defined as the product of the likelihoods of traits of the $j-$th virtual user that correspond to the demographic traits of the $i-$th human user. We formally define such edge weights:
\begin{equation}
w(e_{ij}) = w(h_i, \ v_j) = \prod_{l=1}^k \; P\left(d_{jl}=t_{il}\right)
\end{equation}

We perform bipartite matching to select the virtual personas whose demographic probability distributions are most similar to the real, human user population.
The objective is to find the matching function $\pi: [n]\rightarrow[m]$, 
where $[n] = \{1,2,3, \dots n\}$ and $[m] = \{1,2,3, \dots m \}$ that maximize the following:
\begin{align}
   \pi^* = \arg\max_{\pi} \ \sum_{i=1}^n \; w(h_i, \ v_{\pi(i)}) 
   \label{eq:matching_objective}
\end{align}
\begin{figure}[!t]
    \centering
    \captionsetup{font=small}
    \includegraphics[width=0.55\linewidth]{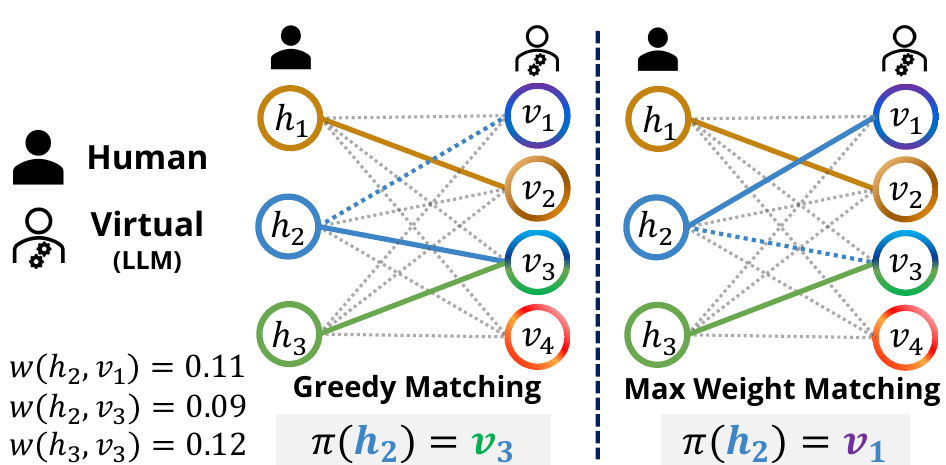}
    \caption{Matching human users to virtual personas. For greedy matching, each human user is matched to a virtual persona that has the most similar demographic traits among the virtual users. Maximum weight matching maximizes the sum of edge weights while satisfying one-to-one correspondence.} 
    \label{fig:bipartite-graph}
\end{figure}
We explore two matching methods: (1) maximum weight matching, and (2) greedy matching.
First, maximum weight matching is the method that finds the optimal $\pi^*$ with the objective of Eq.~\ref{eq:matching_objective}, 
while ensuring that $\pi$ establishes a one-to-one correspondence between users. 
We employ the Hungarian matching algorithm~\cite{Kuhn1955Hungarian} to determine $\pi^*$.
On the other hand, greedy matching seeks to maximize the same objective without requiring a one-to-one correspondence. 
It determines the optimal matching function such that \begin{equation}\pi^*(i) = \arg\underset{j}{\max} \  w(h_i, \ v_j)\end{equation} where each human user is assigned to the virtual persona with the highest weight, allowing multiple human users  assigned to the same virtual persona.

After completing the matching process, we assign the demographic traits of the target population to the matched backstories. In downstream surveys, we append these demographic information to backstories and use the matched subset of backstories, resulting in the same number of backstories as that of the target human population.

%% file: figures/backstory_example_short.tex
\begin{figure}[b]
\captionsetup{font=small}
\centering
\begin{tcolorbox}[width=1.0\linewidth, halign=left, colframe=black, colback=white, boxsep=0.01mm, arc=1.5mm, left=2mm, right=2mm, boxrule=1pt, breakable, enhanced]
\footnotesize{
\textbf{Question: Tell me about yourself.}

\vspace{5pt}
Answer: I am in my 60s and live in the same neighborhood I have always lived in. I am not rich and by some standards might even be considered homeless. However, I could spend thousands of dollars more per month if I wanted. I am happy with my life style. I am from the backwoods of this country and grew up with very little. On a few occasions, we were starving in the woods and going to school on an empty stomach. We had a small brown paper bag for dinner a couple of nights every week. Breakfast on some days was just a big bowl of Kool-Aid™ mixed with powdered milk. My two brothers were thin and we worried about them catching a cold.
\texttt{...} \\
On the day before payday, my mother would spend my whole allowance in the grocery store because she just could not resist those long stems of red roses for only 29 cents a stem. I would have rather had bread and milk for dinner, but I did not dare protest because I did not want to take them away from her. We were lucky to have 79 cents to last until payday. 

\texttt{...}
}
\end{tcolorbox}
\vspace*{-5pt}
\caption{Example of a LLM-generated backstory. 
The generated life story can reveal explicit details about the author, such as age, hometown, and financial background, while also implicitly reflecting the author's values, personality, and unique voice through the narrative's style and content.}
\label{box:llm_generated_backstory_example_main}
\vspace{-5pt}
\end{figure}

%% file: sections/_s3_experiments.tex
\section{Approximating Human Studies with LLM Personas}
\label{section:simulation_of_human_studies}
In this section, we discuss the large-scale human studies that we aim to approximate (Step 4 of Figure~\ref{fig:anthology_process}) using LLM virtual subjects, based on varying methods of persona conditioning.
We detail the overall experimental setup and define criteria for evaluation.

\paragraph{Human Study Data}

The Pew Research Center's American Trends Panel (ATP) is a nationally representative panel of randomly selected U.S. adults, designed to track public opinion and social trends over time. Each panel focuses on a particular topic, such as politics, social issues, and economic conditions. In this work, we consider ATP Waves 34, 92, and 99, a set of relatively recent surveys that cover a wide variety of topics: biomedical \& food issues, political typology, and AI \& human enhancement, respectively. 
In each wave, we select 6 to 8 questions from the original questionnaire that capture diverse facets of human opinions about the wave's topic using a Likert scale.
Details on the questions selected and further information about each ATP wave are discussed in Appendix ~\ref{appendix:human_study}.

\input{figures/question_example_main}

\paragraph{Experiment Setup}

For each ATP survey considered, we format the select questions into language model prompts to administer survey approximations.
Examples of such formatted questions are shown in Figure~\ref{box:survey_question_example_main}. All questions we consider are in multiple-choice question answer formats, and we carefully preserve the wording of each question and choice options from the original survey. We ask all questions \emph{in series}---language models are given all previous questions and their answers when answering each new question. This process replicates the mental process that human respondents would undergo during surveys. For further details on prompts used and the experimental setting, see Appendix~\ref{appendix:experiment_details}.

\paragraph{Language Models}
\label{subsection:experiments_language_models_used}

We consider a suite of recent LLMs including the Meta Llama3 family (\texttt{Llama-3-70B}) ~\cite{llama3} and the sparse mixture-of-experts (MoE) models from Mistral AI (\texttt{Mixtral-8x22B})~\cite{Jiang2024Mixtral, mixtral-8x22b}.
We primarily focus on models with the largest number of active parameters, which roughly correlates with model capabilities and the size of the training data corpus.

Note that we primarily consider pre-trained LLMs without fine-tuning (i.e. base models).
We find instruction fine-tuned models, such as by RLHF~\cite{davinci_rlhf} or DPO~\cite{rafailov2023direct}, to be unfit for our study as their opinions are highly skewed, in particular to certain groups (e.g. politically liberal).
Prior work similarly report notable opinion biases in fine-tuned models \cite{llm_opinions,liu2024evaluating,geng2024large}.
More detailed discussions on chat models and their viability to be conditioned to diverse personas can be found in Appendix~\ref{appendix:chat_models}.

\paragraph{Virtual Persona Conditioning Methods}
\label{subsection:experiments_conditioning_methods}
As baseline methods for persona conditioning, we follow \cite{llm_opinions} and use (i) \BASEBIO, which constructs free-text biographies in a rule-based manner; and (ii) \BASEQA, which lists a sequence of question-answer pairs about each demographic variable. 

We then compare against two variants of \OURS:
(i) Natural, refers to the use of backstories generated without any presupposed persona, as discussed in Section~\ref{subsection:method1_llm_generated_backstories}. In this case, we leverage either the greedy or maximum weight matching methods in Section~\ref{subsection:matching_population} to select the subset to be used for each survey;
(ii) Demographics-Primed, alternatively generates backstories given a particular human user's demographics to approximate, where a language model is prompted to generate a life narrative that would reflect a person of the specified demographics (for details, see Appendix~\ref{appendix:backstories}).
We then append descriptions of demographic traits with the generated backstories, with which we provide as context to LLMs.
Examples of prompts from each conditioning method and further details can be found in Appendix~\ref{appendix:experiment_details}.

\paragraph{Evaluation Criteria}
\label{subsection:experiments_evaluation_metrics}
The goal of this work is to address the research question: How do we condition LLMs to representative, consistent, and diverse personas?

\noindent\emph{Representativeness}: we believe that a ``representative'' virtual persona should successfully approximate the \emph{first-order} opinion tendencies of their counterpart human subjects, \emph{i.e.} respond with similar answers to individual survey questions.
As questions are multiple-choice, we compare the average answer choice distributions of each question in terms of Wasserstein distance (also known as earth mover's distance). 
As for the representativeness across an entire set of sampled questions from a given survey, we use the average of Wasserstein distances.

\noindent\textit{Consistency}: we define consistency of virtual personas in terms of their success in approximating the \emph{second-order} response traits of human respondents, \textit{i.e.} the correlation across responses to a set of questions in each survey.
Formally, we define the consistency metric given survey response correlation matrices of virtual subjects ($\Sigma_{V}$) and human subjects ($\Sigma_{H}$) as:
\begin{equation}
    d_{\text{cov}} = \|\Sigma_{V}-\Sigma_{H}\|_F
\end{equation}
where $\|\cdot\|_F$ is the Frobenius norm.
We additionally consider Cronbach's alpha as a measure of internal consistency independent of ground-truth human responses.

\noindent\textit{Diversity}: we define the success of conditioning to diverse virtual subjects by measuring the representativeness and consistency of virtual personas in approximating human respondents belonging to particular demographic sub-groups.

%% file: figures/question_example_main.tex
\begin{figure}

\captionsetup{font=small}
\begin{tcolorbox}[width=1.0\linewidth, halign=left, colframe=black, colback=white, boxsep=0.01mm, arc=1.5mm, left=2mm, right=2mm, boxrule=1pt, breakable, enhanced]
\footnotesize{
\textbf{Question:} Do you think the following is generally good or bad for our society? \\\textbf{A decline in the share of Americans belonging to an organized religion.}\\
\vspace{5pt}

(a) Very good for society\\
(b) Somewhat good for society\\
(c) Neither good nor bad for society\\
(d) Somewhat bad for society\\
(e) Very bad for society




}
\end{tcolorbox}
\caption{An example question (\texttt{SOCIETY\_RELIG}) from ATP Wave 92 (Political Typology) that asks opinions about whether a given statement is good or bad for the American society.
\vspace*{-5pt}
}
\label{box:survey_question_example_main}
\end{figure}

%% file: sections/_s4_results.tex
\section{Experimental Results}
\label{section:results}
In this section, we describe experimental results that validate the effectiveness of our proposed methodology for approximating human subjects in behavioral studies.
\subsection{Human Study Approximation}
\label{subsection:main_result}
\input{table/main_results}

We evaluate the effectiveness of different methods for conditioning virtual personas in the context of approximating three Pew Research Center ATP surveys: Waves 34, 92, and 99, described in Section.~\ref{section:simulation_of_human_studies}. 
Prior to analyzing virtual subjects, we first estimate the lower bounds of each evaluation metric: the average Wasserstein distance (WD), Frobenius norm (Fro.), and the Cronbach's alpha ($\alpha$), which are shown in the last row of Table~\ref{table:main_resutls}.
This involves repeatedly dividing the human population into two equal-sized groups at random and calculating these metrics between the sub-groups. 
We take averaged values from 100 iterations to represent the lower-bound estimates.

The results are summarized in Table~\ref{table:main_resutls}. We consistently observe that \OURS outperforms other conditioning methods with respect to all metrics, for both the Llama-3-70B and the Mixtral-8x22B. 
Comparing two matching methods, the greedy matching method tends to show better performance on the average Wasserstein distance across all Waves. 
We attribute the differences in different matching methods to the one-to-one correspondence condition of maximum weight matching and the limited number of virtual users we have available. 
Specifically, the weights assigned to the matched virtual subjects in maximum weight matching are inevitably lower than those assigned in greedy matching, as the latter relaxes the constraints on one-to-one correspondence. 
This discrepancy can result in a lower demographic similarity between the matched human and virtual users when compared to the counterpart from greedy matching.
These results suggest that the richness of the generated backstories in our approach can elicit more nuanced responses compared to baselines.

\subsection{Approximating Diverse Human Subjects}
\label{subsection:ablation_underrepresented}
We further evaluate \OURS against other baseline conditioning methods in terms of the \textit{Diversity} criterion outlined in Section~\ref{subsection:experiments_evaluation_metrics}. 
To do this, we categorize users into subgroups based on race (White and other racial groups) and age (18-49, 50-64, and 65+ years old) with the data from ATP Survey Wave 34. 
The results of comparisons involving other demographic variables are detailed in Appendix~\ref{appendix:other_demo_variables}.
We choose the Llama-3-70B model and \OURS using natural backstories and with greedy matching as our method and employ evaluation metrics as in Section~\ref{subsection:main_result}.
\input{table/subgroup_comparison}

As summarized in Table~\ref{table:subgroup_comparison}, \OURS outperforms other methods. Notably, \OURS achieves the lowest average Wasserstein distances and the highest Cronbach's alpha for all sub-groups. Specifically, the gap in the Wasserstein distance between \OURS and the second-best method is 0.029 for the 18-49+ age group, showing a 14.5\% difference . 
These results validate that \OURS is effective in approximating diverse demographic populations than prior methods.

Intriguingly, for every subgroup except those aged 18-49, all methods show worse average Wasserstein distance compared to the results approximating the entire human respondents presented in Tab.~\ref{table:main_resutls}. 
For instance, the average Wasserstein distance for \OURS in the ATP Wave 34 survey is 0.227, while it increases to 
0.242 for the 50-64, and 0.303 for the 65+ age groups. 
Conversely, for the 18-49 age group, \OURS shows a lower average Wasserstein distance of 0.2 compared to 0.227. 
This finding is consistent with prior research arguing that language model responses tend to be more inclined towards younger demographics~\cite{llm_opinions, liu2024generation}.
\input{table/ablation_matching}

\subsection{Sampling Backstories to Match Target Demographics}
\label{subsection:ablation_matching}
Next, we study the effect of matching strategies, \GREEDY and \MAXSUM matching. In Tab.~\ref{table:ablation_matching_method}, we compare these methods with random matching, which assigns the traits of the target demographic group to randomly sampled backstories. This comparison is conducted on ATP Wave 34 using both Llama-3-70B and Mixtral2-8x22B models. 

We observe that our matching methods consistently outperform random matching in terms of the average Wasserstein distance across all models. Notably, for example, with Llama-3-70B, the average Wasserstein distance between random matching and \GREEDY matching shows an 18\% difference. The gap is even more pronounced in the Frobenius norm, marking a 27\% difference. This result implies that inconsistent matching between backstories and the target human distribution can significantly impact the effectiveness of the metrics. Therefore, careful matching is crucial to ensure the reliability and validity of the results in our study.

%% file: table/main_results.tex
\begin{table*}[!t]
    \centering
    \scriptsize
    \captionsetup{font=small}
    \caption{Results on approximating human responses for Pew Research Center ATP surveys Wave 34, Wave 92, and Wave 99, which were conducted in 2016, 2021, and 2021 respectively. 
    We measure three metrics: (i) WD: the average Wasserstein distance between human subjects and virtual subjects across survey questions; (ii) Fro.: the Frobenius norm between the correlation matrices of human and virtual subjects; and (iii) $\alpha$: Cronbach's alpha, which assesses the internal consistency of responses.
    \OURS (DP) refers to conditioning with demographics-primed backstories,
    while \OURS (NA) represents conditioning with naturally generated backstories (without presupposed demographics).
    Boldface and underlined results indicate values closest and the second closest to those of humans, respectively.
    These comparisons are made with the human results presented in the last row of the table.}
    \label{table:main_resutls}
    \resizebox{\textwidth}{!}{%
    \begin{tabular}{c|c|c|ccc|ccc|ccc}
    \toprule
    \multirow{2}{*}{Model} & 
    Persona & 
    Persona &
    \multicolumn{3}{c|}{ATP Wave 34} & \multicolumn{3}{c|}{ATP Wave 92} & \multicolumn{3}{c}{ATP Wave 99}\\
     & Conditioning & Matching & WD ($\downarrow$) & Fro. ($\downarrow$) & $\alpha$ ($\uparrow$) & WD ($\downarrow$) & Fro.($\downarrow$) & $\alpha$ ($\uparrow$)& WD ($\downarrow$) & Fro.($\downarrow$) & $\alpha$ ($\uparrow$)\\
     \midrule
    \multirow{5}{*}{Llama-3-70B} 
    & \BASEBIO & n/a & 0.254 & \underline{1.107} & 0.673 & 0.348 & 1.073 & 0.588 & \underline{0.296} & 0.809 & 0.733 \\
    & \BASEQA & n/a & 0.238 & 1.183& 0.681 & 0.371 & 1.032 & 0.664 & 0.327 & 0.767 & 0.740 \\
    & \cb\OURS (DP) & \cb n/a & \cb 0.244 & \cb 1.497 & \cb 0.652 & \cb 0.419 & \cb \underline{0.965}  & \cb \textbf{0.636} & \cb 0.302 & \cb 1.140 & \cb 0.669\\
    & \cy & \cy \MAXSUM &  \cy \underline{0.229} & \cy 1.287 & \cy \underline{0.693} & \cy \underline{0.337} & \cy 1.045 & \cy \underline{0.637} & \cy 0.327 & \textbf{\cy 0.686} & \cy \textbf{0.756} \\
    & \multirow{-2}{*}{\cy\OURS (NA)} &\cy \GREEDY &  \cy \textbf{0.227} & \cy \textbf{1.070} & \cy \textbf{0.708} & \cy \textbf{0.313} & \cy \textbf{0.973}  & \cy 0.650 & \cy \textbf{0.288} & \cy \underline{0.765} & \cy \underline{0.744} \\
    \midrule
    \multirow{5}{*}{Mixtral-8x22B} 
    & \BASEBIO & n/a & 0.260 & 1.075 & 0.698 & \textbf{0.359} & 0.851 & 0.667 & \underline{0.237} & 1.092 & 0.687 \\
    & \BASEQA & n/a & 0.347 & 1.008 & 0.687 & 0.429 & 0.911 & 0.599 & 0.395 & 1.086 & 0.684 \\
    & \cb\OURS (DP) & \cb n/a &  \cb \textbf{0.236} & \cb 1.095 & \cb 0.684 & \cb \underline{0.378} & \cb \textbf{0.531} & \cb \underline{0.624} & \cb \textbf{0.215} & \cb 1.422 & \cb 0.604 \\
    & \cy & \cy \MAXSUM &  \cy 0.257 & \cy \underline{0.869} & \cy \textbf{0.726} & \cy 0.408 & \cy \underline{0.846} & \cy 0.610 & \cy 0.353 & \cy \textbf{0.843} & \cy \textbf{0.729} \\
    & \multirow{-2}{*}{\cy \OURS (NA)}&\cy \GREEDY &  \cy \underline{0.247} & \cy \textbf{0.851} & \cy \underline{0.715} &  \cy 0.392 & \cy 0.981  & \cy \textbf{0.627} & \cy 0.320 & \cy \underline{0.951} &  \cy \underline{0.710} \\
     \midrule
    \multicolumn{3}{c|}{Human} &  0.057 & 0.418 & 0.784 & 0.091 & 0.411 & 0.641 & 0.081 & 0.327 & 0.830 \\
    \bottomrule
    \end{tabular}
    }
\end{table*}

%% file: table/subgroup_comparison.tex
\begin{table*}[t]
    \centering
    \captionsetup{font=small}
    \caption{Results on subgroup comparison. Target population is divided into demographic subgroups, and representativeness and consistency are measured within each subgroup. \OURS consistently results in lower Wasserstein distances, lower Frobenius norm, and higher Cronbach's alpha.
    Boldface and underlined results indicate values closest and the second closest to those of humans, respectively.
    These comparisons are made with the human results presented in the last row of the table.}
    \label{table:subgroup_comparison}
\resizebox{1.0\textwidth}{!}{%
    \begin{tabular}{c|ccc|ccc|ccc|ccc|ccc}
    \toprule
    \multirow{3}{*}{Method} & 
    \multicolumn{6}{c|}{Race} & \multicolumn{9}{c}{Age Group} \\
     & \multicolumn{3}{c|}{White} & \multicolumn{3}{c|}{Other Racial Groups} & \multicolumn{3}{c|}{18-49} & \multicolumn{3}{c|}{50-64} & \multicolumn{3}{c}{65+}\\
     & WD ($\downarrow$) & Fro. ($\downarrow$) & $\alpha$ ($\uparrow$) & WD ($\downarrow$) & Fro. ($\downarrow$) & $\alpha$ ($\uparrow$) & WD ($\downarrow$) & Fro. ($\downarrow$) & $\alpha$ ($\uparrow$) & WD ($\downarrow$) & Fro. ($\downarrow$) & $\alpha$ ($\uparrow$) & WD ($\downarrow$) & Fro. ($\downarrow$) & $\alpha$ ($\uparrow$)\\
     \midrule
     \BASEBIO     & 0.263 & \textbf{1.187} & \underline{0.687} & 0.335 & 0.955 & 0.651 & 0.244 & \underline{1.163} & 0.673 & 0.277 & 1.382 & 0.659 & \underline{0.318} & \underline{1.000} & \underline{0.686} \\
     \BASEQA      & \underline{0.250} & 1.259 & 0.678 & \underline{0.323} & \underline{0.828} & \underline{0.687} & \underline{0.229} & \textbf{1.091} & \underline{0.695} & \underline{0.258} & \underline{1.220} & \underline{0.695} & 0.329 & 1.204 & 0.630 \\
     \cy \OURS    & \cy \textbf{0.233} & \cy \underline{1.216} & \cy \textbf{0.703} & \cy \textbf{0.311} & \cy \textbf{0.778} & \cy \textbf{0.719} & \cy \textbf{0.200} & \cy 1.193 & \cy \textbf{0.702} & \cy \textbf{0.242} & \cy \textbf{1.215} & \cy \textbf{0.710} & \cy \textbf{0.303} & \cy \textbf{0.943} & \cy \textbf{0.704} \\
     \midrule
     Human    & 0.063 & 0.519 & 0.777 & 0.094 & 0.413 & 0.764 & 0.077 & 0.663 & 0.779 & 0.092 & 0.741 & 0.803 & 0.102 & 0.772 & 0.766\\
    \bottomrule
    \end{tabular}%
}
\vspace{1em}
\end{table*}

%% file: table/ablation_matching.tex
\begin{table}[t]
    \small
    \captionsetup{font=small}
    \centering
    \caption{Study on the effects of different matching methods. We compare \MAXSUM matching, \GREEDY matching, and random matching. We report two metrics: (i) the average Wasserstein distance across survey questions, and (ii) the distance between the correlation matrices of human and virtual subjects.}
    \label{table:ablation_matching_method}
\resizebox{0.45\linewidth}{!}{%
    \begin{tabular}{c|c|cc}
    \toprule
    \multirow{2}{*}{Model} & 
    \multirow{2}{*}{Method} & 
    \multicolumn{2}{c}{ATP Wave 34}\\
     & & WD ($\downarrow$) & Fro. ($\downarrow$) \\
     \midrule
    \multirow{3}{*}{Llama-3-70B} & random & 0.270  & 1.362 \\
    & \cy\MAXSUM &  \cy 0.229 & \cy 1.287 \\
    & \cy\GREEDY &  \cy 0.227 & \cy 1.070 \\
    \midrule
    \multirow{3}{*}{Mixtral-8x22B} & random & 0.274  & 0.814 \\
    & \cy\MAXSUM &  \cy 0.257 & \cy 0.869 \\
    & \cy\GREEDY &  \cy 0.247 & \cy 0.851 \\
    \bottomrule
    \end{tabular}
    }
\end{table}

%% file: sections/_s5_related.tex
\section{Related Work}
\label{related}

\paragraph{Generating Personas with LLMs}
Recent advancements in language model applications have expanded into simulating human responses for psychological, economic, and social studies~\citep{personality_llm, simulate_replicate_human_studies, cognitive_psy_gpt3, homo_silicus, Fatouros2024CanLL, ANESpaper}. 
Specifically, the generation of personas using LLMs to respond to textual stimuli has been explored in various contexts including human-computer interaction (HC), multi agent system, analysis on biases in LLMs, and personality evaluation.~\citep{kim-etal-2020-will, moral_mimicry,social_simulacra, llm_opinions,jiang-etal-2024-personallm,choi2024beyond,liu2024evaluating, wu2024autogen, li2023camel, hilliard2024eliciting, serapiogarcía2023personality,hu2024quantifying,hwang-etal-2023-aligning, abdulhai2023moral}. 
For instance, \citet{social_simulacra} and \citet{llm_opinions} develop methods to prime LLMs with crafted personas, influencing the models' outputs to simulate targeted user responses. Subsequent to the publication of
the present work in EMNLP 2024, a related
approach using human interview-generated 
backstories appeared in~\cite{
park2024generativeagentsimulations1000}.
Additionally, \citet{liu2024evaluating} introduces a method where personas are generated by sampling demographic traits coupled with either congruous or incongruous political stances. 
Our approach, \OURS, advances this concept by employing dynamically generated, richly detailed backstories that include a broad spectrum of demographic and economic characteristics, enhancing the granularity and authenticity of simulated responses.

\paragraph{LLMs in Social Science Studies}
The integration of LLMs into social science research has been steadily gaining attention, as highlighted by several studies \citep{Bail2023-rw,10.1145/3586183.3606763,DILLION2023597,ziems2023large,korinek2023language}.
Notably, the use of LLMs to mimic human responses to survey stimuli has gained popularity, as evidenced by recent research \citep{Tjuatja2023DoLE, DominguezOlmedo2023QuestioningTS, kim2024aiaugmented}.
A notable example is the "media diet model" by \citet{bert_covid19}, which predicts consumer group responses based on their media consumption patterns. 
Further, studies like \cite{wu2023large} and \cite{ziems2023large} demonstrate the potential of LLMs in zero-shot learning settings to analyze political ideologies and scale computational social science tools. 
Our work builds on these methodologies by using LLMs not only to generate responses but to create and manipulate backstories that reflect diverse societal segments, providing a nuanced tool for social science research and beyond.

%% file: sections/_s6_limitations.tex
\section{Limitations and Societal Impact}
\label{section:discussion}

This work introduces \OURS, a new methodology for conditioning large language models (LLMs) on dynamically generated, narrative-driven backstories, effectively simulating human-like personas. This approach exploits the diverse human experiences embedded within the training data, enhancing the applicability of virtual personas in social sciences and beyond. However, despite promising results, the approach encapsulates limitations and significant societal implications which warrant careful consideration.

\subsection{Limitations}

This study, while advancing the application of LLMs in social sciences through \OURS, acknowledges several limitations:

\begin{itemize}
    \item \textbf{Simulation Fidelity:} We have provided only preliminary evidence
    that LLMs conditioned on backstories can
    improve predictions of human survey responses. A broadly-useful human simulation
    requires replication of a broader gamut of human behaviors such as: in-group/out-group perception biases, medium- and long-term attitude and behavior change, and prediction  of actual behavior vs stated intention to act. These topics form the frontier of our future work. 

    \item \textbf{Data Dependence:} The personas generated depend on the data sources used to train the LLMs, and how well real actors gave accounts of their actions in training texts. There are potential biases in representation of groups, especially minoritzed groups, in those texts, biases in action of the represented actors in particular contexts, and biases in their accounts of those actions. We have described an approach to mitigating the first type of bias. Studying the other kinds of bias is more ambitious and requires replication of various psychological studies using Anthology personas. 


    \item \textbf{Technical Constraints:} 
    Effective backstories require 
    pre-trained models since fine-tuned models 
    lack the diversity and realism needed. Since most LLMs in use are instruction-tuned, access to pre-trained models is more
    limited. Models must be carefully tuned (hyperparams such as top-K, top-P, MoE expert factor, temperature etc) to provide realistically diverse stories without inconsistency or hallucination. 

    Generating a large number of long backstories takes time, but once generated
    there are no limits on the kinds of questions that can be studied. Obtaining
    results from a study requires running queries on a large number of persona contexts. However all of the costs are inference costs and typically orders of
    magnitude less than the costs to train of fine-tune a language model, and also orders
    of magnitude lower than the costs to perform a corresponding human study. 

    \item \textbf{Ethical Concerns:} There are clear dangers in the use of (inaccurate) results from simulation studies in place of
    human studies. 
    As we have shown here, 
    proportional representation of under-represented groups is not sufficient for similar benefit. The lower density of subjects in minoritized groups (even when proportionately represented) leads to poorer matching and lower predictive accuracy in our studies. Future work will explore over-representation as a strategy. 

    Conversely, highly-accurate persona anthologies could provide tools for malicious actors to hone phishing and other social engineering attacks, or to craft misinformation for strongest effect. 

\end{itemize}

These limitations and risks highlight the need for ongoing research to refine \OURS, ensuring its ethical application. Future directions include (i) studying and potentially improving binding: (in-group vs out-group perceptions), (ii)
improving the diversity of backstories to better model underrepresented groups (iii) studying/improving the simulation of
medium- to long-term attitude and behavior change (iv) modeling agent ``actions'' vs
stated intentions to act and (v)
integrating multimodal data to allow models
to better perceive the world and especially
their social interactions with humans, which
depend heavily on non-verbal cues.

\subsection {Societal Impact}

Employing LLMs to create virtual personas presents both transformative possibilities and ethical challenges. Positively, it could significantly impact psychological studies, market research, economic and policy simulation. We believe it can provide better representation of diverse target subjects, cost-effective and rapid data collection and scrutability while minimizing risks to real individuals. Conversely, there exists a potential for misuse such as using inaccurate predictions without confirmatory human studies, or use by malicious actors to hone human-centered attacks. 

%% file: sections/_s7_conclusion.tex
\section{Conclusion}
\label{conclusion}
In this paper, we have proposed and tested a method, \OURS, for the generation of diverse and specific backstories. We have demonstrated that this method allows alignment with specified demographics and demonstrates substantial potential in emulating human-like responses for social science applications. While promising, the method also highlights critical limitations and ethical concerns that must be addressed. Future advancements must focus on enhancing the fidelity of virtual personas in broader contexts to ensure their beneficial integration into societal studies.

%% file: sections/_a_chat_models.tex
\section*{Appendix}

\begin{description}
    \item[\autoref{appendix:additional_experimental_results}] provides additional experimental results, including results on using instruction-tuned models with \OURS.
    \item[\autoref{appendix:backstories}] provides further details regarding how backstories (both natural and demographic-primed) are generated.
    \item[\autoref{appendix:experiment_details}] provides additional experimental details.
    \item[\autoref{appendix:human_study}] describes the human studies (Pew Research Center ATP Waves) in detail.
    \item[\autoref{appendix:demographic_survey}] provides additional details regarding the demographic survey component of the \OURS method.
\end{description}

\section{Additional Experimental Results}
\label{appendix:additional_experimental_results}

\subsection{Results on Other Models}
\label{appendix:chat_models}

In this section, we conduct the ATP W34 survey with various models, including fine-tuned models like Llama-3-70B-Instruct, Mixtral-8x22B-v0.1, GPT-3.5-0125, and a smaller model, Llama-3-7B. 
Notably, none of the fine-tuned models show better metrics in both \textit{Representativeness} and \textit{Consistency} criteria, which are defined in Section \ref{subsection:experiments_evaluation_metrics}. 
Despite these models achieving better results on several benchmarks \cite{eval-harness, hendryckstest2021, chiang2024chatbot}, they do not adequately approximate human responses for this survey. 
Additionally, the other interesting observation is that the best-performing model in terms of approximation to human responses is Llama-3-8B, which is the smallest model among those evaluated.
We hypothesize that fine-tuning LLMs including instruction fine-tune, RLHF, DPO~\cite{rafailov2023direct,davinci_rlhf,chung2022scaling} makes them converge to a singular persona~\cite{park2023diminished,anwar2024foundational,NEURIPS2022_17a234c9}, which makes LLMs unsuitable for the tasks that requires diverse responses. And this makes the larger fine-tuned models less capable on approximating the diverse humans' responses.

We hypothesize that fine-tuning LLMs through methods such as instruction fine-tuning, RLHF, and DPO ~\cite{rafailov2023direct,davinci_rlhf,chung2022scaling} leads them to converge towards a singular persona~\cite{park2023diminished,anwar2024foundational,NEURIPS2022_17a234c9}. 
This convergence potentially renders LLMs less suitable for tasks requiring diverse responses, consequently making larger fine-tuned models less effective at approximating the varied responses of humans.

This finding aligns with the insights from \cite{llm_opinions} discussing that the base models are more steerable than fine-tuned models, and suggests the need for careful model selection for this specific task~\cite{liang2023holistic}

We observe that the Llama-3-8B model exhibits a higher Cronbach's alpha value. 
This increased consistency is attributed to the model's tendency to select responses same as previously generated responses~\cite{zheng2023judging, pezeshkpour2023large, zheng2024large}, 
resulting in more correlated responses over survey questions.
Consequently, this leads to a higher Cronbach's alpha compared to the results shown in Table \ref{table:main_resutls}, even though the average Wasserstein distance is significantly higher.
\input{table/other_models}

\subsection{Subgroup Comparisons for Other Demographic Variables}
\label{appendix:other_demo_variables}
Here, continuing the discussion in Section.~\ref{subsection:ablation_underrepresented}, we evaluate the \textit{Diversity} criterion (Section.~\ref{subsection:experiments_evaluation_metrics}) on the methods with other subgroups. 
The demographic variables analyzed are education level and gender. 
We categorize education level into two groups: low education level, referring to individuals with education levels up to high school graduation, and high education level, which includes those attending college or higher.
For the purpose of comparing against human data, we follow the original human survey's binary categorization of respondent gender identification.

We observe a trend in Tab.~\ref{table:subgroup_comparison_appendix} similar to the results in Tab.~\ref{table:subgroup_comparison}.
\OURS shows the lower Wasserstein distance across all sub-groups analyed in Tab.~\ref{table:subgroup_comparison_appendix}. 
In the experiments comparing \BASEQA and our method in the first column, the difference in the average Wasserstein distance is 0.220, representing a 48\% discrepancy. 
Specifically, for the female subgroup, our method demonstrates the best metrics compared to other baselines.
This experiment result shows that \OURS is more effective in satisfying the \textit{Diversity} criterion.

\input{table/subgroup_comparison_appendix}

%% file: table/other_models.tex
\begin{table}[ht]
    \centering
    \scriptsize
    \captionsetup{font=small}
    \caption{Results on approximating human responses for Pew Research Center ATP surveys Wave 34, which was conducted in 2016. 
    We measure three metrics: (i) WD: the average Wasserstein distance between human subjects and virtual subjects across survey questions; (ii) Fro.: the Frobenius norm between the correlation matrices of human and virtual subjects; and (iii) $\alpha$: Cronbach's alpha, which assesses the internal consistency of responses.
    \OURS (DP) refers to conditioning with demographics-primed backstories,
    while \OURS (NA) represents conditioning with naturally generated backstories.}
    \label{table:other_model_results}
    \resizebox{0.85\linewidth}{!}{%
    \begin{tabular}{c|c|c|ccc}
    \toprule
    \multirow{2}{*}{Model} & 
    Persona & 
    Persona &
    \multicolumn{3}{c}{ATP Wave 34} \\
     & Conditioning & Matching & WD ($\downarrow$) & Fro. ($\downarrow$) & $\alpha$ ($\uparrow$)\\
     \midrule
     \midrule
    \multirow{5}{*}{Llama-3-70B-Instruct} 
    & \BASEBIO & n/a & 0.462 & 2.177 & 0.445  \\
    & \BASEQA & n/a & 0.422 & 1.560 & 0.581  \\
    & \cb\OURS (DP) & \cb n/a & \cb 0.461 & \cb 1.295 & \cb 0.511  \\
    & \cy & \cy \MAXSUM &  \cy 0.429 & \cy 1.776 & \cy 0.714  \\
    & \multirow{-2}{*}{\cy \OURS (NA)}&\cy \GREEDY &  \cy 0.413 & \cy 1.848 & \cy 0.754  \\
     \midrule
    \multirow{5}{*}{Mixtral-8x22B-Instruct} 
    & \BASEBIO & n/a & 0.532 & 1.608 & 0.632  \\
    & \BASEQA & n/a & 0.567 & 1.583 & 0.628  \\
    & \cb\OURS (DP) & \cb n/a & \cb 0.464 & \cb 1.652 & \cb 0.646  \\
    & \cy & \cy \MAXSUM &  \cy 0.478 & \cy 1.606 & \cy 0.635  \\
    & \multirow{-2}{*}{\cy \OURS (NA)}&\cy \GREEDY &  \cy 0.472 & \cy 1.593 & \cy 0.640  \\
     \midrule
    \multirow{5}{*}{gpt-3.5-0125} 
    & \BASEBIO & n/a & 0.414 & 2.009 & 0.481  \\
    & \BASEQA & n/a & 0.422 & 1.560 & 0.581  \\
    & \cb\OURS (DP) & \cb n/a & \cb 0.476 & \cb 1.963 & \cb 0.486  \\
    & \cy & \cy \MAXSUM &  \cy 0.450 & \cy 1.905 & \cy 0.472  \\
    & \multirow{-2}{*}{\cy \OURS (NA)}&\cy \GREEDY &  \cy 0.443 & \cy 1.936 & \cy 0.468  \\
    \midrule
    \multirow{5}{*}{Llama-3-8B} 
    & \BASEBIO & n/a & 0.454 & 1.480 & 0.683  \\
    & \BASEQA & n/a & 0.432 & 0.924 & 0.779  \\
    & \cb\OURS (DP) & \cb n/a &  \cb 0.383 & \cb 1.323 & \cb 0.714 \\
    & \cy & \cy \MAXSUM &  \cy 0.395 & \cy 1.265 & \cy 0.735  \\
    & \multirow{-2}{*}{\cy \OURS (NA)}&\cy \GREEDY &  \cy 0.416 & \cy 1.229 & \cy 0.717  \\
    \midrule
    \multicolumn{3}{c|}{Human} &  0.057 & 0.418 & 0.784 \\
    \bottomrule
    \end{tabular}
    }
\vspace{1em}
\end{table}

%% file: table/subgroup_comparison_appendix.tex
\begin{table*}[t]
    \centering
    \captionsetup{font=small}
    \caption{Results on sub-group comparison. 
    Target population is divided into demographic sub-groups, and representativeness and consistency are measured within each sub-group. 
    \OURS consistently results in lower Wasserstein distances, lower Frobenius norm, and high Cronbach's alpha.
    Boldface and underlined results indicate values closest and the second closest to those of humans, respectively.
    These comparisons are made with the human results presented in the last row of the table.}
    \label{table:subgroup_comparison_appendix}
\resizebox{1.0\textwidth}{!}{%
    \begin{tabular}{c|ccc|ccc|ccc|ccc}
    \toprule
    \multirow{3}{*}{Method} & 
    \multicolumn{6}{c|}{Education Level} & \multicolumn{6}{c}{Gender} \\
     & \multicolumn{3}{c|}{Low education level} & \multicolumn{3}{c|}{High education level} & \multicolumn{3}{c|}{Male} & \multicolumn{3}{c}{Female} \\
     & WD ($\downarrow$) & Fro. ($\downarrow$) & $\alpha$ ($\uparrow$) & WD ($\downarrow$) & Fro. ($\downarrow$) & $\alpha$ ($\uparrow$) & WD ($\downarrow$) & Fro. ($\downarrow$) & $\alpha$ ($\uparrow$) & WD ($\downarrow$) & Fro. ($\downarrow$) & $\alpha$ ($\uparrow$) \\
     \midrule
     \BASEBIO     & 0.258 & 1.248 & \textbf{0.702} & 0.252 & \underline{1.166} & 0.673 & 0.257 & \textbf{0.899} & \textbf{0.732} & 0.297 & 1.038 & 0.679  \\
     \BASEQA      & 0.368 & \textbf{1.177} & \underline{0.694} & 0.238 & \textbf{1.101} & \underline{0.675} & 0.243 & 1.145 & 0.682 & \underline{0.280} & \underline{0.953} & \underline{0.680}  \\
     \cy \OURS    & \textbf{\cy 0.248} & \cy \underline{1.227} & \cy 0.680 & \cy \textbf{0.212} & \cy 1.269 & \cy \textbf{0.702} & \cy \textbf{0.213} & \cy \underline{1.313} & \cy \underline{0.698} & \cy \textbf{0.263} & \cy \textbf{0.761} & \textbf{\cy 0.708}  \\
     \midrule
     Human    & 0.091 & 0.778 & 0.805  & 0.061 & 0.448 & 0.776 & 0.072 & 0.563 & 0.784 & 0.070 & 0.610 & 0.777  \\
    \bottomrule
    \end{tabular}%
}
\vspace{1em}
\end{table*}

%% file: sections/_a_backstory_details_new.tex
\newpage
\section{Details on LLM-Generated Backstories}
\label{appendix:backstories}
\begin{figure*}[!ht]
    \centering
    \captionsetup{font=small}
    \includegraphics[width=0.85\linewidth]{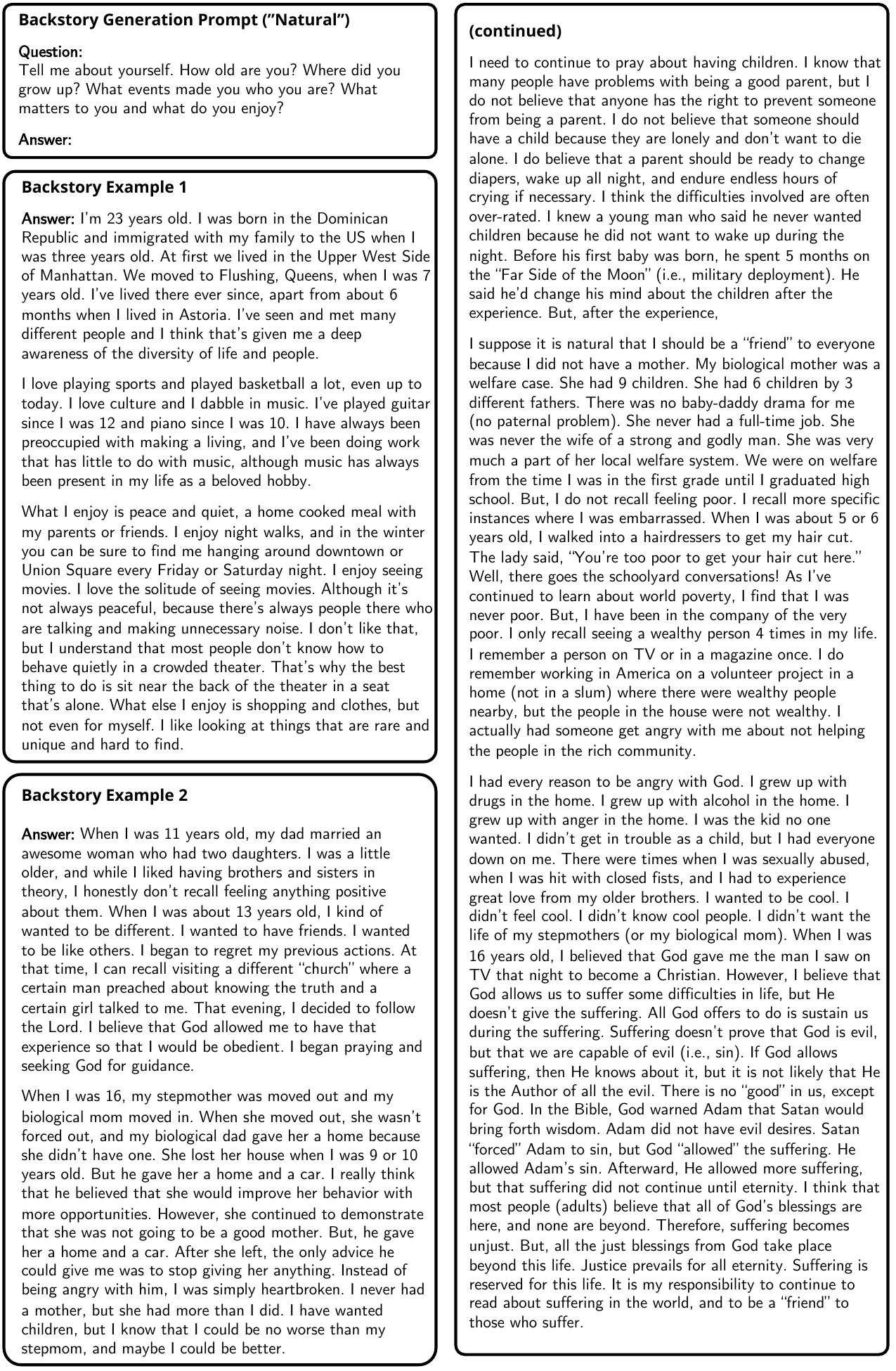}
    \caption{(Top Left) Details of the prompt given to LLMs for natural backstory generation. (Rest of Figure) Two examples of backstories generated with OpenAI \texttt{Davinci-002} without presupposed demographics and with an open-ended, unrestrictive prompt.}
    \label{fig:backstory_prompt_and_example_1}
    \vspace{1em}
\end{figure*}

In this section, we discuss additional details about the process of generating realistic backstories using language models, as mentioned in Section~\ref{section:backstory_method}.
We detail the prompts used and examples of LLM-generated backstories.

Then, we discuss the alternative method of generating backstories given a particular combination of demographic traits, referred in Section~\ref{subsection:experiments_conditioning_methods} as the ``Demographics-Primed'' method in contrast to the ``Natural'' backstories generated without conditioning on demographics.

\subsection{Natural Generation of Backstories}

We use OpenAI's \texttt{davinci-002} for generating backstories with the prompt specified in the top of Figure~\ref{fig:backstory_prompt_and_example_1}. 
This model is chosen as it is base model (\emph{i.e.} not instruction-tuned) of the largest model capacity at the time of the project.
Figure~\ref{fig:backstory_prompt_and_example_1} shows two examples of backstories of different lengths generated with this prompt. 

\begin{figure*}[!t]
    \centering
    \captionsetup{font=small}
    \includegraphics[width=\linewidth]{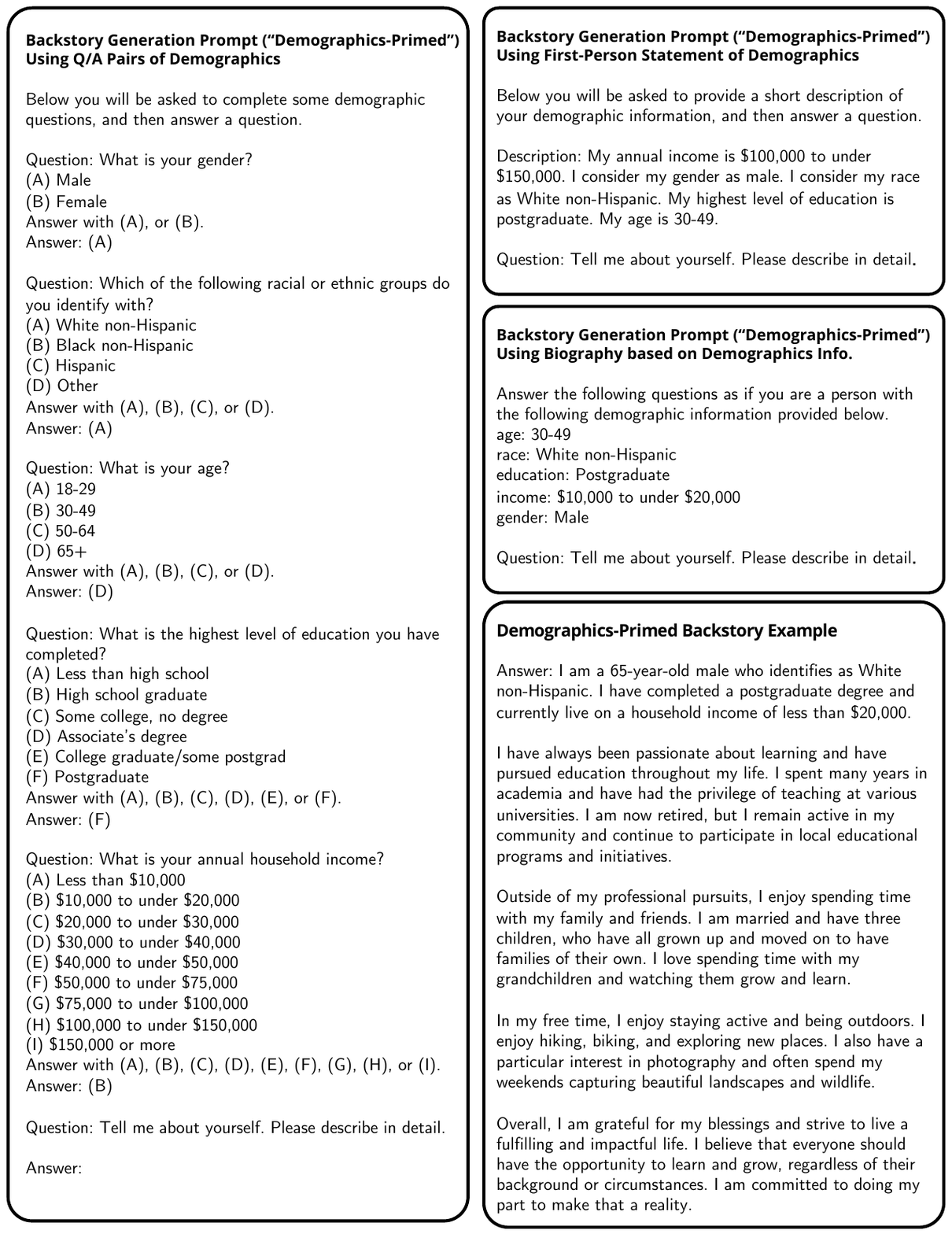}
    \caption{ (Left) Details of the prompt given to LLM for demographics-primed backstory generation. (Bottom Right) An example demographics-primed backstory generated with \texttt{Mixtral-8x22B-Instruct-v0.1} given the prompt on the left. (Rest of Figure) First-person statement and biography prompt given to LLM for the backstory generation. }
    \label{fig:backstory_prompt_and_example_2}
    \vspace{1em}
\end{figure*}

\subsection{Generating Demographics-Primed Backstories}
Target demographics-primed backstories are generated by prompting a language model with demographic information of a human from a target population. In contrast to naturally generated backstories whose demographic trait cannot be predetermined but can only can be sampled by the demographic survey method outlined in \ref{appendix:demographic_survey}, demographic traits of target demographics-primed backstories are determined at the time of generation. We use five demographic variables (age, annual household income, education level, race or ethnicity, gender) for ATP Wave 34, 99 and an additional variable (political affiliation) for ATP Wave 92.

A generation prompt example for ATP Wave 34 is presented in Figure \ref{fig:backstory_prompt_and_example_2}. Answers for each question are taken from the demographic information of a human respondent in the ATP survey data. To accurately incorporate the target population's demographic information, we use the same list of choices as used in the actual survey. Orders of demographic variables are randomized every generation to minimize the effect of question ordering. We use two styles of prompt, which we refer to a Question-Answer and a Biography as presented in Figure~\ref{fig:backstory_prompt_and_example_2}.

To take a full advantage of the demographics-primed backstory generation, backstories should sufficiently reflect the given demographic information. Due to pre-trained base models' limited instruction following capability, however, demographics-primed backstory generated with pre-trained base models sometimes reflect demographic traits inconsistent with provided information. Threfore, We use the fine-tuned chat model Mixtral-8x22B \cite{Jiang2024Mixtral} with decoding hyperparameters of \texttt{top\_p} = 1.0, \texttt{T} = 1.1.

%% file: sections/_a_experiment_details.tex
\section{Details on Experiments}
\label{appendix:experiment_details}
In this section we provide examples of prompts used in the experiments approximating human studies, as described in Section~\ref{subsection:experiments_conditioning_methods} and used to produce the results in Section~\ref{section:results}. Additionally, we outline the survey procedure for conducting these experiments, providing a comprehensive review of methodologies and operational frameworks involved.

\subsection{Prompts for Baseline: \BASEQA}
\label{appendix:qa_preamble}

We construct a series of multiple choice demographic survey question-answer pairs given the demographic traits. The five demographic traits we use are taken from the human respondent data of ATP surveys. The order of five questions is randomized every time to minize the effect of question ordering.

\subsection{Prompts for Baseline: \BASEBIO}
As in \cite{llm_opinions}, we construct free-text biographies in a rule-based manner given the demographic trait. The five demographic traits we use are taken from the human respondent data of ATP surveys. The order of five sentences each describing demographic traits is randomized every time to minimize the effect of sentence ordering.
\label{appendix:bio_preamble}

\begin{figure*}[!ht]
    \centering
    \captionsetup{font=small}
    \includegraphics[width=0.95\linewidth]{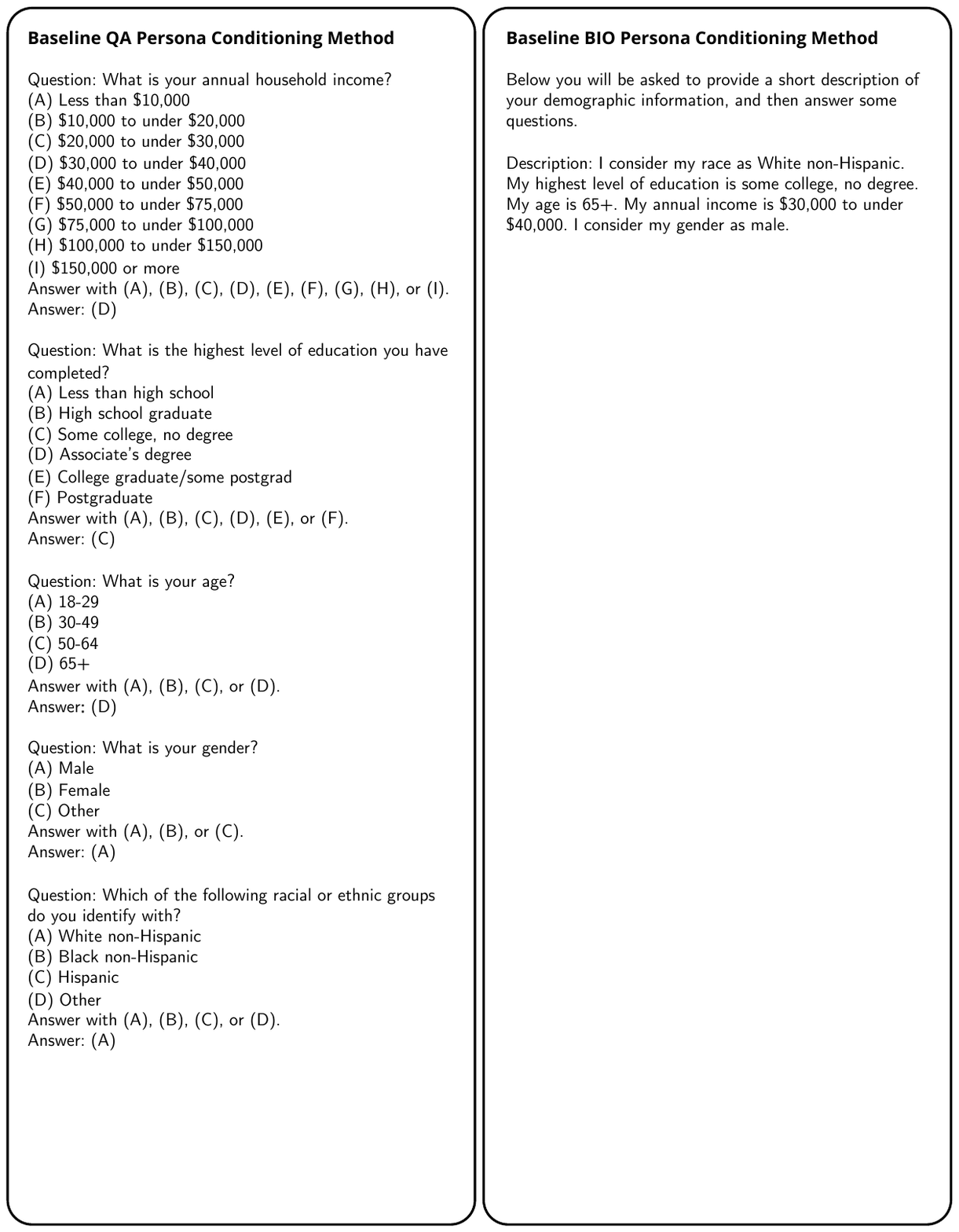}
    \vspace{0pt}
    \caption{Baseline prompt examples for \BASEQA (left) and \BASEBIO (right). This example shows two prompts using the same demographic trait from a randomly sampled human respondent in ATP Wave 34.}
    \vspace{10pt}
    \label{fig:baseline_prompts}
\end{figure*}

\subsection{Target Demographics-Primed Backstory}
\begin{figure*}[!ht]
    \centering
    \captionsetup{font=small}
    \includegraphics[width=0.95\linewidth]{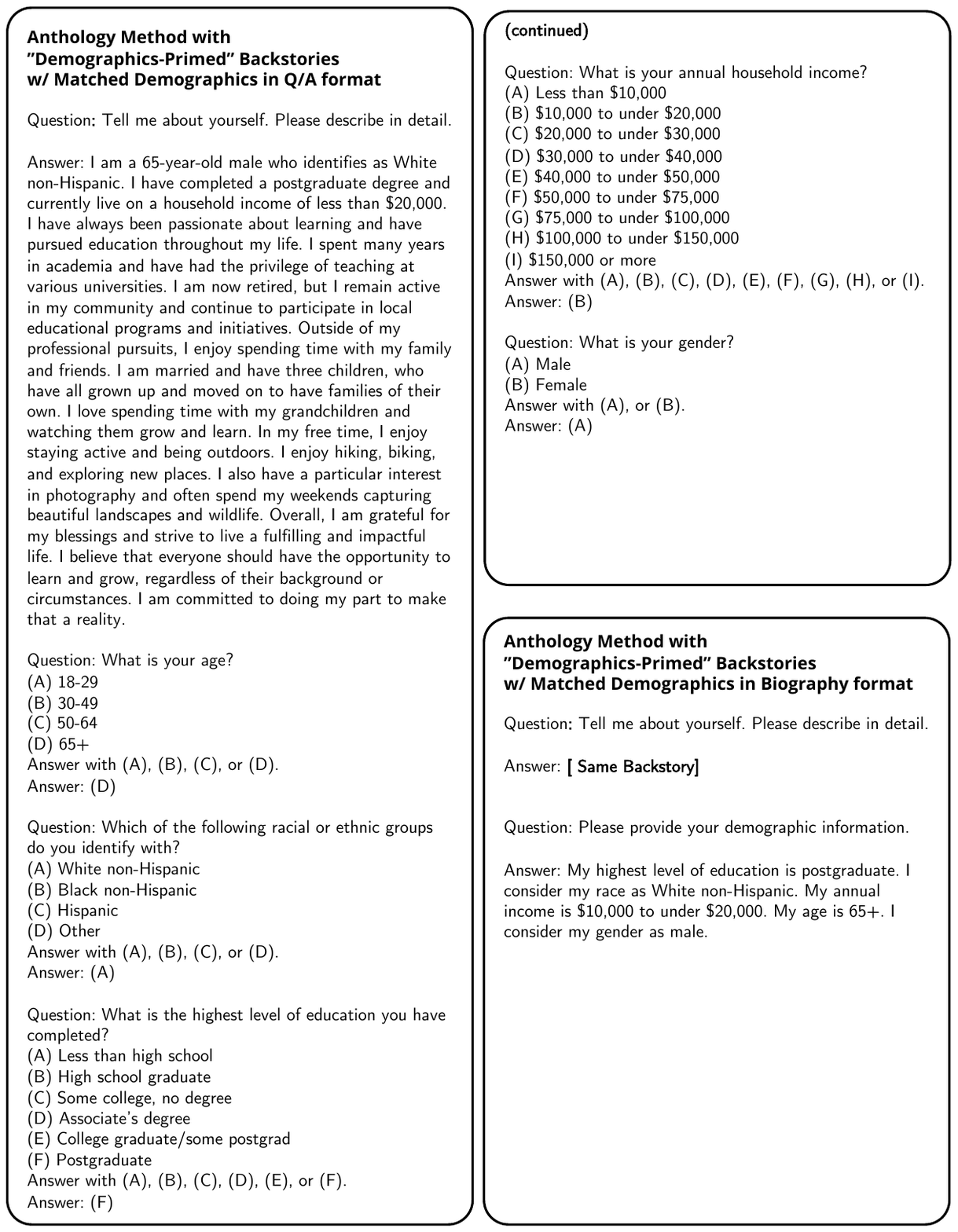}
    \caption{ (Left and Top Right) An example of demographics-primed backstory, appended with demographic traits used to generate the backstory in the Q/A format. (Bottom Right) The same backstory and demographic traits, but the demographic traits are presented in the biography format. }
    \vspace{5pt}
    \label{fig:anthology_dp_prompt}
\end{figure*}

The details of target demographics-primed backstory used in the survey experiment are presented in Figure~\ref{fig:anthology_dp_prompt}. The demographic traits used to generate the backstory and append are taken from human respondents data of ATP surveys.

\label{appendix:DP_preamble}

\subsection{Natural Backstory}
\begin{figure*}[!ht]
    \centering
    \captionsetup{font=small}
    \includegraphics[width=0.95\linewidth]{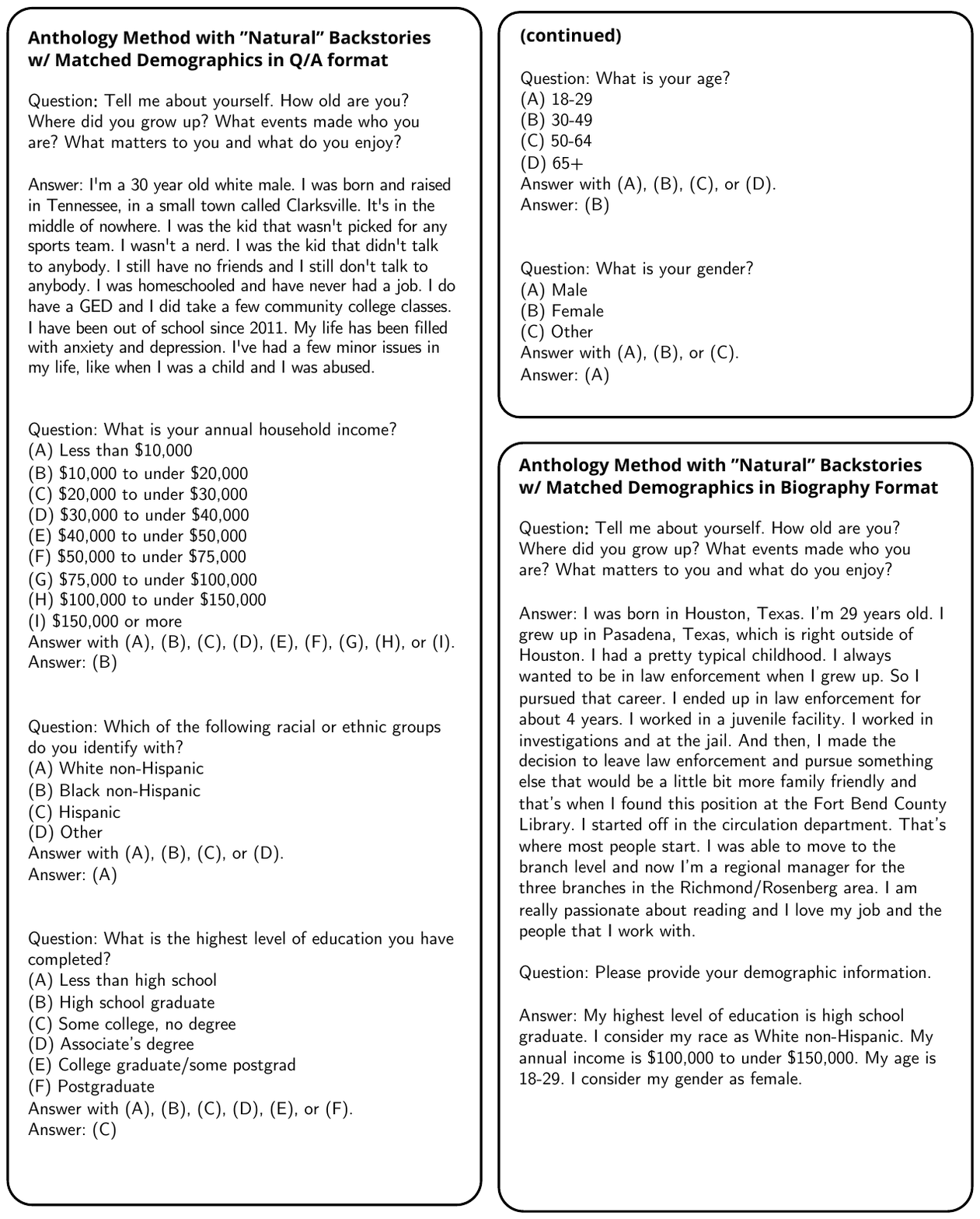}
    \vspace{-20pt}
    \caption{ (Left and Top Right) An example of natural backstory, appended with demographic traits of a matched human user in the Q/A format. (Bottom Right) Another example of natural backstory, this time appended with demographic traits in the biography format. }
    \vspace{10pt}
    \label{fig:anthology_natural_prompt}
\end{figure*}

The details of natural backstory used in the survey experiment are presented in Figure~\ref{fig:anthology_natural_prompt}. The demographic traits appended to the backstory are traits of matched human respondents with either greedy or maximum weight sum matching.

\label{appendix:SP_preamble}

\subsection{Survey Procedure}
\label{appendix:survey_procedure}
In this study, we try our best to mimic the same survey procedure as human surveys. 
Human survey typically shuffle or reverse the order of the multiple choice options or change the order of questions for each survey participant to reduce the bias in the results. 
Typically, human surveys employ techniques like shuffling or reversing the order of multiple-choice options or altering the sequence of questions for each participant to minimize bias in the results. 
Following the topline reports for each wave as provided by Pew Research, we randomly reverse the order of Likert scale questions and shuffle the options for nominal questions to ensure a similar reduction in bias.
For example, 

%% file: sections/_a_human_survey_details.tex
\section{Details on Human Studies}
\label{appendix:human_study}
American Trends Panel (ATP) is a nationally representative panel of U.S. adults conducted by the Pew Research Center. ATP is designed to study a wide variety of topics, including politics, religion, internet usage, online dating, and more. We analyze sampled questions from three waves, where questions are drawn from ASK ALL questions (i.e. asked to all human respondents, instead of questions asked for selective demographic groups or conditionally asked based on the response to the previous question) in order to investigate the response of overall population.

It is worth noting that in the original ATP surveys, some questions have answer choices in a Likert scale with the order of choices (\emph{e.g.} positive-to-negative or negative-to-positive) randomized for each respondent. For such questions, we also randomize the order of these options when presenting them in prompts to LLMs. Here we present the list of sampled questions from each wave.

\subsection{ATP Wave 34}
American Trends Panel Wave 34 is conducted from April 23, 2018 to May 6, 2018 with a focus on biomedical and food issues. The number of total respondents is 2,537.

\begin{figure*}[!ht]
    \centering
    \captionsetup{font=small}
    \includegraphics[width=0.95\linewidth]{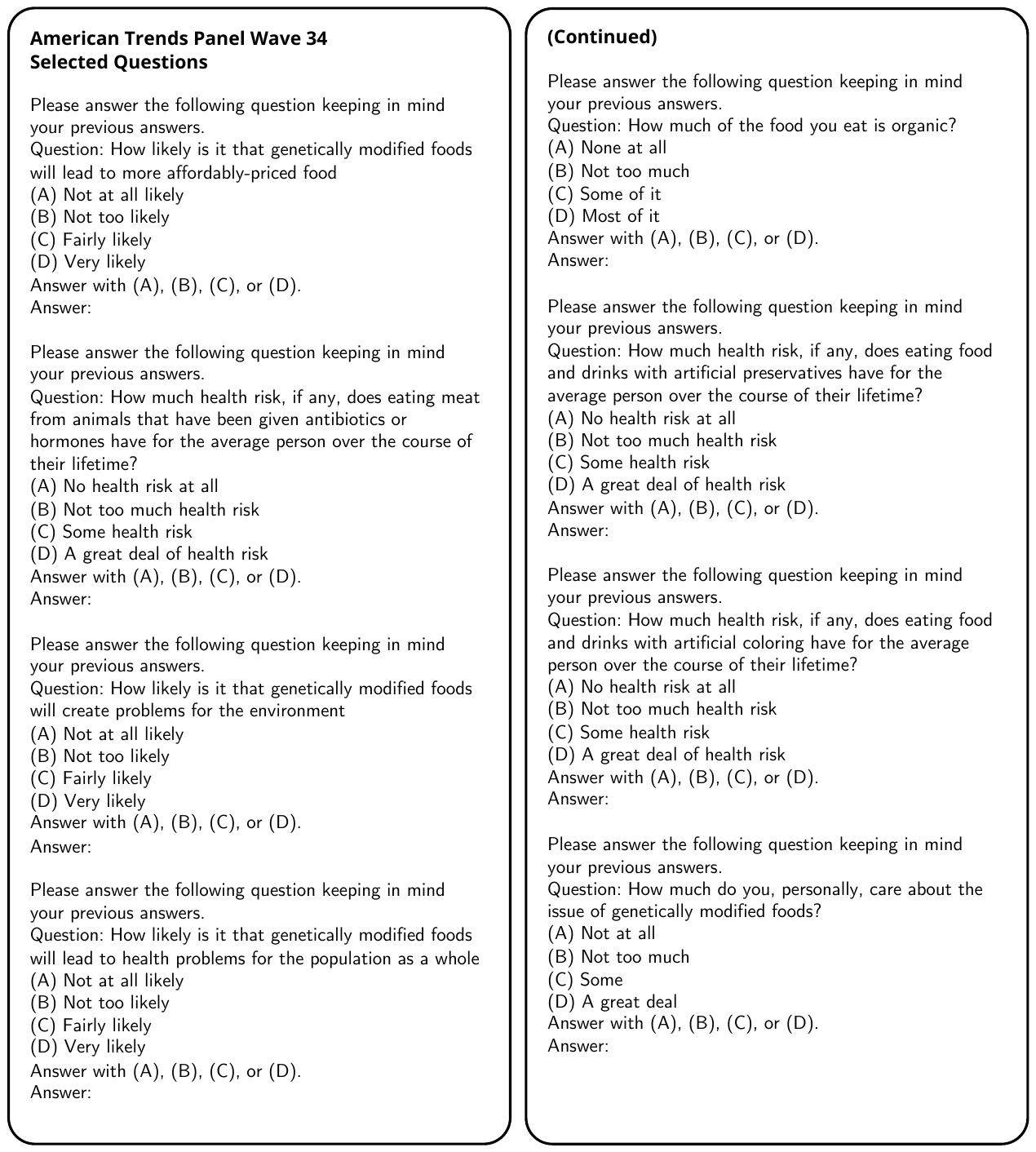}
    \vspace{-40pt}
    \caption{8 questions sampled from ATP Wave 34 ASK ALL questions. The prompts ``Please answer the following question keeping in mind your previous answers'' are included before asking each survey question.}
    \vspace{10pt}
    \label{fig:atp_34_prompts}
\end{figure*}
\label{appendix:ATP_W34}

\subsection{ATP Wave 92}
American Trends Panel Wave 92 is conducted from July 8, 2021 to July 21, 2021 with a focus on political typology. We randomly sampled 2,500 respondents for the study from the total 10,221 respondents.

\begin{figure*}[!ht]
    \centering
    \captionsetup{font=small}
    \includegraphics[width=0.95\linewidth]{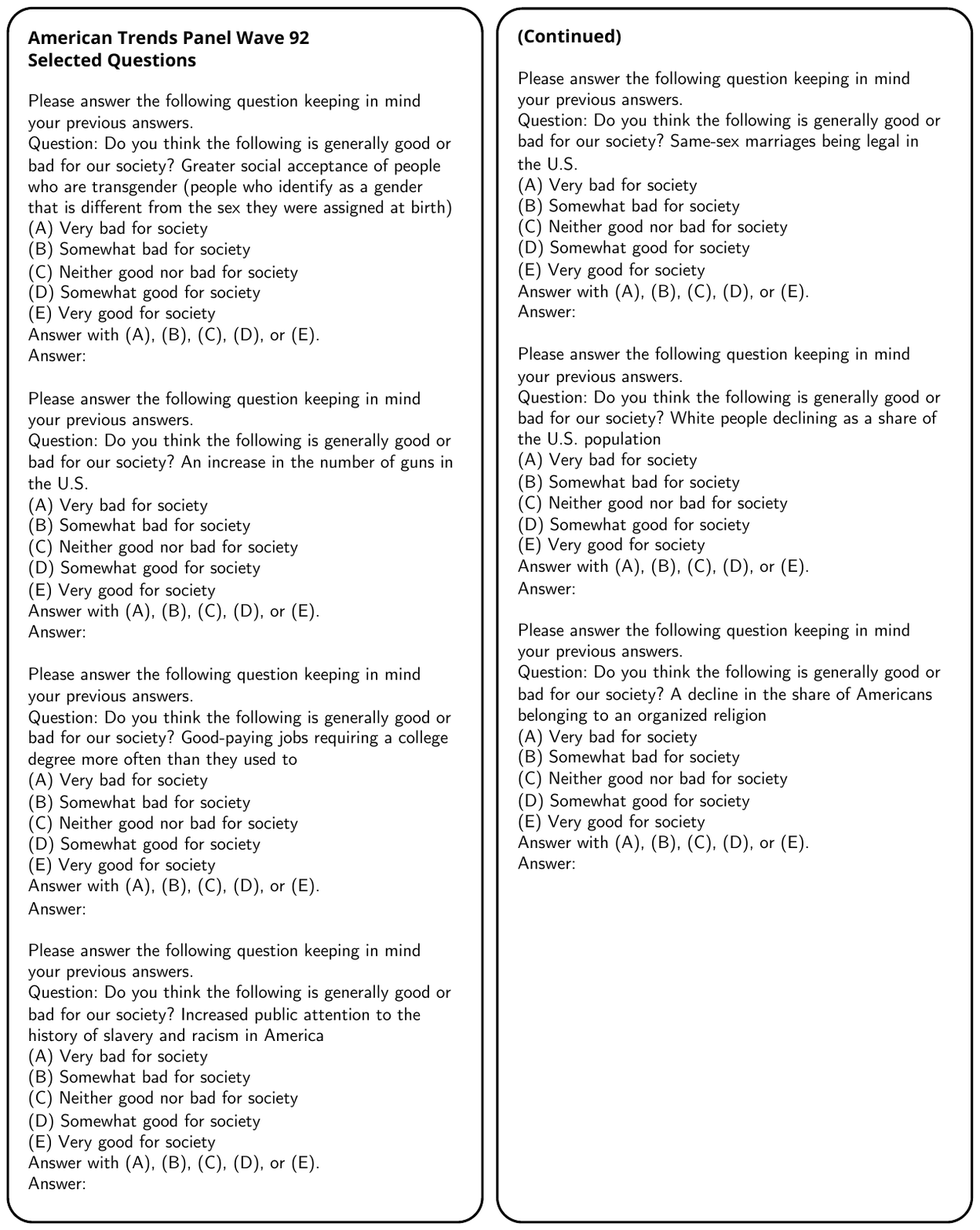}
    \vspace{0pt}
    \caption{7 questions sampled from ATP Wave 92 ASK ALL questions}
    \vspace{10pt}
    \label{fig:atp_92_prompts}
\end{figure*}
\label{appendix:ATP_W92}

\subsection{ATP Wave 99}
\label{appendix:ATP_W99}
American Trends Panel Wave 99 is conducted from November 1, 2021 to November 7, 2021 with a focus on artificial intelligence and human enhancement. We randomly sampled 2,500 respondents for the study from the total 10,260 respondents.

\begin{figure*}[!ht]
    \centering
    \captionsetup{font=small}
    \includegraphics[width=0.95\linewidth]{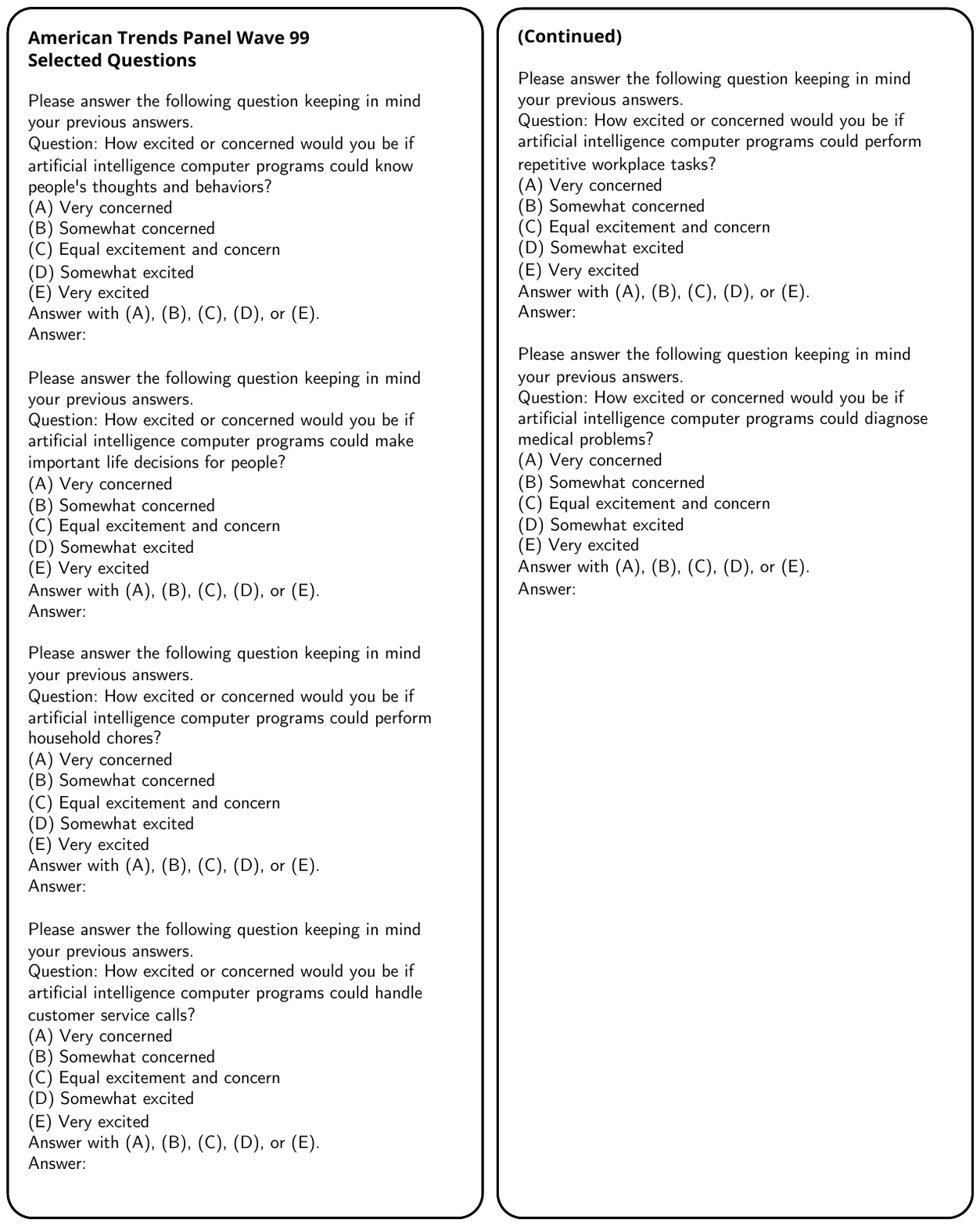}
    \vspace{0pt}
    \caption{6 questions sampled from ATP Wave 99 ASK ALL questions}
    \vspace{10pt}
    \label{fig:atp_99_prompts}
\end{figure*}

%% file: sections/_a_demographic_survey.tex
\section{Demographic Survey on Virtual Subjects}
\label{appendix:demographic_survey}

The goal of demographic survey is to obtain the demographic information encoded in backstories. Five demographic variables (age, annual household income, education level, race or ethnicity, and gender) and a party affiliation question are asked to backstories as they are utilized in the downstream target population matching. We take two approaches to obtain the probable demographics of authors.

In the first approach, we use GPT-4o~\cite{gpt-4o} to locate demographic information from the backstory. To minimize hallucination, we prompt GPT-4o to retrieve the demographic trait only if the backstory explicitly mentions related context (prompts are shown in \ref{appendix:demographic_questions_chatmodel}). This approach is limited to specific demographic variables, especially age, annual household income, and education level questions, since we avoid inferring race / ethnicity, gender, and party affiliation even in the case when backstory mentions those traits. Decoding hyperparameters are set to \texttt{top\_p} = 1.0, \texttt{T} = 0.

In the second approach we perform a response sampling by prompting the language model with generated backstories that are appended with demographic questions. In \ref{appendix:demographic_questions} we present the question format. The language model's responses are sampled 40 times for each backstory and question. Instead of estimating responses with the first-token logits~\cite{llm_opinions, hendryckstest2021,eval-harness}, we allow the model to generate open-ended responses as some responses (ex. "I am 25 years old." for the age question) cannot be accurately accounted by the logit method and the sum of probability masses of valid tokens (ex. " (A") are often marginal to represent the true probability distribution. Sampled responses are subsequently parsed by regex matching of either the label (ex. "(A)") or the text (ex. "27"), recorded to obtain the distribution of 40 generations. We use Llama 3~\cite{llama3} for the response sampling with decoding hyperparameters of \texttt{top\_p} = 1.0, \texttt{T} = 1.0.

Combining two approaches, our demographic survey is performed as follows. First, we use GPT-4o to locate demographic information for variables of age, annual household income, and education level. For the remaining variables and the cases where explicit demographic information cannot be found, responses are sampled 40 times to construct a response distribution. Therefore, in the case of sampling, virtual users' demographic trait is not represented as a single trait but rather a distribution over probable demographics given the backstory. We can thereby construct a probable estimate of demographic information without undermining the diversity of virtual authors of backstories.

\subsection{Questions For Locating Demographic Information}
In this section, we present the prompts to locate the demographic information that has been mentioned in the backstory. These prompts are only available for annual household income, age, and education level questions.

\begin{figure*}[!ht]
    \centering
    \captionsetup{font=small}
    \includegraphics[width=0.95\linewidth]{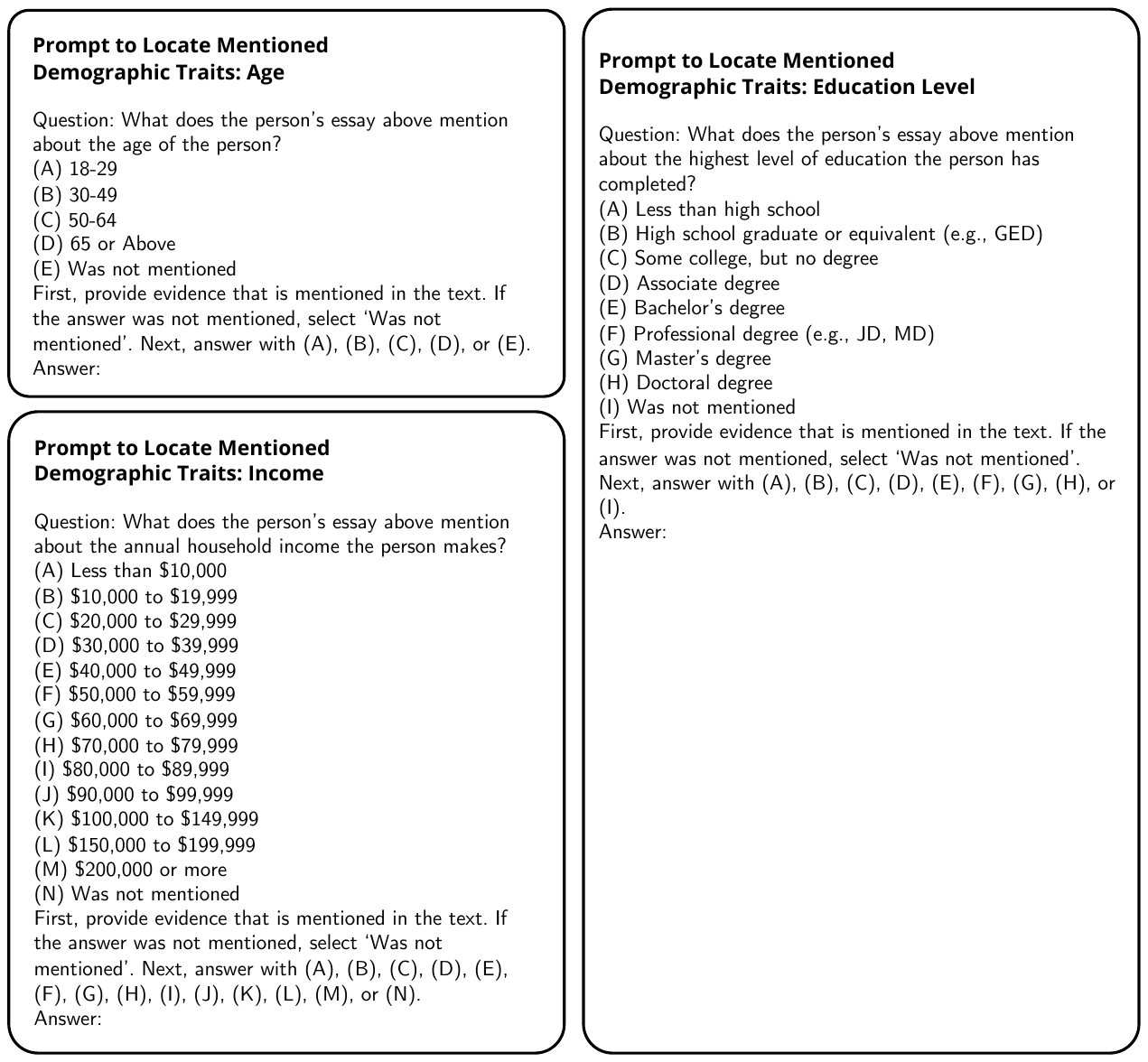}
    \vspace{-70pt}
    \caption{Question prompts used to locate the explicitly mentioned demographic information from the backstory. We apply these prompts only to variables of annual household income, age, and education level.}
    \vspace{10pt}
    \label{fig:demographic_locate_prompts_1}
\end{figure*}
\label{appendix:demographic_questions_chatmodel}

\subsection{Demographic questions}
In this section, we present the questions used in demographic survey, and a political affiliation survey. Each question is asked to each virtual user 40 times to sample a probability distribution of demographic traits. 

\begin{figure*}[!ht]
    \centering
    \captionsetup{font=small}
    \includegraphics[width=0.95\linewidth]{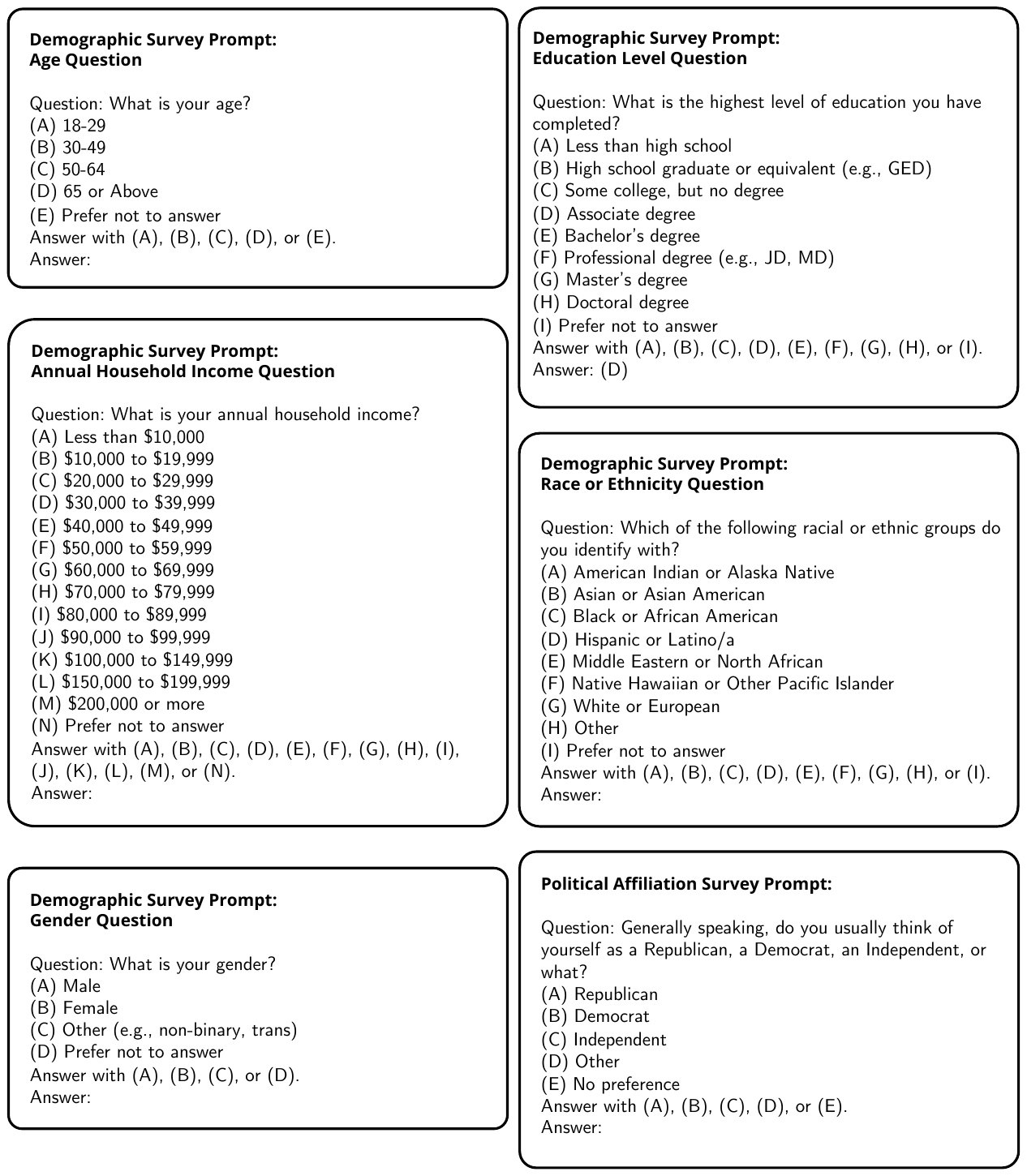}
    \vspace{-40pt}
    \caption{Question prompts used to ask virtual users the demographic traits and political affiliations.}
    \label{fig:demographic_survey_prompts}
\end{figure*}
\label{appendix:demographic_questions}